\documentclass[11pt]{article}

\usepackage[margin=1in]{geometry}
\usepackage{amsthm,amsmath,amsfonts,amssymb}
\usepackage[authoryear]{natbib}
\usepackage[colorlinks,citecolor=blue,urlcolor=blue,linkcolor=blue,bookmarksnumbered]{hyperref}
\usepackage{bookmark}
\usepackage{graphicx}
\usepackage{cleveref}
\usepackage{algorithm}
\usepackage{algorithmic}

\usepackage{setspace}
\usepackage{authblk}

\makeatletter
\let\old@paragraph\paragraph
\def\paragraph{\@ifstar\sada@paragraph@s\sada@paragraph@n}
\def\sada@paragraph@s#1{\old@paragraph*{#1.}}
\def\sada@paragraph@n#1{\old@paragraph{#1.}}
\makeatother

\theoremstyle{plain}

\newtheorem{theorem}{Theorem}[section]
\newtheorem{lemma}[theorem]{Lemma}
\newtheorem{proposition}{Proposition}

\theoremstyle{definition}

\newtheorem{remark}{Remark}
\newtheorem{assumption}{Assumption}
\usepackage{bm}
\usepackage{booktabs}
\usepackage{xcolor}
\usepackage{tcolorbox}
\tcbuselibrary{breakable}
\usepackage{multirow}
\usepackage{graphicx}
\usepackage{subcaption}

\def\calF{\mathcal{F}}

\def\calL{\mathcal{L}}

\def\calN{\mathcal{N}}
\def\calP{\mathcal{P}}

\def\calR{\mathcal{R}}
\def\calS{\mathcal{S}}
\def\calT{\mathcal{T}}
\def\calU{\mathcal{U}}

\def\calW{\mathcal{W}}
\def\calX{\mathcal{X}}
\def\calY{\mathcal{Y}}

\def\0{\mathbf{0}}
\def\A{\mathbf{A}}
\def\B{\mathbf{B}}
\def\H{\mathbf{H}}
\def\I{\mathbf{I}}

\def\U{\mathbf{U}}
\def\V{\mathbf{V}}

\def\X{\mathbf{X}}
\def\Y{\mathbf{Y}}

\def\a{\mathbf{a}}
\def\b{\mathbf{b}}

\def\u{\mathbf{u}}
\def\f{\mathbf{f}}
\def\x{\mathbf{x}}
\def\y{\mathbf{y}}
\def\s{\mathbf{s}}
\def\btt{{\bm\theta}}

\def\bSig{{\bm\Sigma}}

\newcommand{\convd}{\xrightarrow{d}}

\newcommand{\convp}{\xrightarrow{p}}

\def\mR{\mathbb{R}}
\def\mP{\mathbb{P}}

\def\mE{\mathbb{E}}
\def\eif{\text{\normalfont eif}}
\def\ppi{\text{\normalfont ppi}}

\def\opt{\text{\normalfont opt}}

\def\sada{\text{\normalfont sada}}
\def\nv{\text{\normalfont nv}}

\DeclareMathOperator{\cov}{cov}
\DeclareMathOperator{\var}{var}
\DeclareMathOperator{\sd}{sd}

\newcommand{\T}{\text{\scalebox{0.9}{$\top$}}}
\newcommand{\trans}{^\T} 

\def\wh{\widehat}
\def\wt{\widetilde}

\newcommand{\amin}{\operatornamewithlimits{arg\,min}}

\def\boxit#1{\vbox{\hrule\hbox{\vrule\kern6pt\vbox{\kern6pt#1\kern6pt}\kern6pt\vrule}\hrule}}

\title{SADA: Safe and Adaptive Aggregation of Multiple Black-Box Predictions in Semi-Supervised Learning}

\author[1,2]{Jiawei Shan}
\author[2]{Zhifeng Chen}
\author[1]{Yiming Dong}
\author[2]{Yazhen Wang}
\author[1,2]{Jiwei Zhao\thanks{E-mail: \texttt{jiwei.zhao@wisc.edu}}}

\affil[1]{Department of Biostatistics \& Medical Informatics, University of Wisconsin-Madison}
\affil[2]{Department of Statistics, University of Wisconsin-Madison}

\date{}

\begin{document}

\maketitle

\begin{abstract}
  Semi-supervised learning (SSL) arises in practice when labeled data are scarce or expensive to obtain, while large quantities of unlabeled data are readily available. With the growing adoption of machine learning techniques, it has become increasingly feasible to generate multiple predicted labels using a variety of models and algorithms, including deep learning, large language models, and generative AI. In this paper, we propose a novel approach that safely and adaptively aggregates multiple black-box predictions of uncertain quality for both inference and prediction tasks. Our method provides two key guarantees: (i) it never performs worse than using the labeled data alone, regardless of the quality of the predictions; and (ii) if any one of the predictions (without knowing which one) perfectly fits the ground truth, the algorithm adaptively exploits this to achieve either a faster convergence rate or the semiparametric efficiency bound. We demonstrate the effectiveness of the proposed algorithm through small-scale simulations and two real-data analyses with distinct scientific goals. A user-friendly R package, \texttt{sada}, is provided to facilitate practical implementation.
\end{abstract}


\medskip

\section{Introduction}
In many scientific and industrial applications, acquiring gold-standard labels can be expensive, time-consuming, or require specialized expertise, while collecting unlabeled data is relatively cheap and scalable. This disparity motivates semi-supervised learning (SSL), a paradigm that combines scarce labeled observations with large amounts of unlabeled samples \citep{Zhu2005,vanEngelenHoos2020}. Over the past decades, a wide range of SSL algorithms have been developed to exploit abundant unlabeled data to improve model performance, for example by using a pre-trained model to assign pseudo labels to unlabeled samples \citep{Lee2013,ZhuDingJacobsonWuEtAl2023} or by propagating information through relationships between data points \citep{JebaraWangChang2009,SubramanyaTalukdar2014}.

With the rapid evolution of machine learning and generative AI, it has become much easier than ever to generate predicted labels using powerful tools such as large language models (LLMs). These predictions can often be obtained at substantially lower cost and with much shorter turnaround time than human labels. Consequently, leveraging machine-learned predictions in SSL has become an increasingly important practical demand. In the context of statistical inference, \citet{AngelopoulosBatesFannjiangJordanEtAl2023} introduced prediction-powered inference (PPI), a framework that enables valid inference without imposing any assumptions on the predictions.  However, when the predictive model is of low quality, PPI may have worse estimation efficiency than labeled-only baselines. This limitation has motivated further work aimed at improving the efficiency or integrating it with ideas from other statistical and machine learning frameworks \citep{AngelopoulosDuchiZrnic2024,ZrnicCandes2024,FischMaynezHoferDhingraEtAl2024,JiLeiZrnic2025,MiaoMiaoWuZhaoEtAl2025}.

Despite these advances, several gaps and challenges remain when one considers modern SSL pipelines in which multiple predicted labels are available. 
First, current PPI methods are designed for a single predictive model, whereas in practice it is common to obtain predictions from multiple models or sources. Although one may heuristically select the seemingly best model or apply ensembling strategies, there is no general framework within SSL for aggregating multiple predictions in a principled manner.  
Second, outputs from different models, such as GPT, Llama, or DeepSeek, can differ substantially, and the quality of predictions from black box models can be highly variable. In particular, low-quality or poorly calibrated predictions can introduce significant noise, increasing variance and leading to unreliable inference.  
Third, most existing work targets statistical inference for the parameter of interest; however,
in many applications such as image classification, practitioners are primarily concerned with predictive model accuracy rather than parameter estimation. Current PPI methods and their extensions do not directly address how to improve prediction accuracy while retaining rigorous guarantees.

In this paper, we introduce SADA, a novel and comprehensive procedure that safely and adaptively aggregates multiple predicted labels originating from diverse black-box models for both inference and prediction tasks. Our approach is characterized by two key features: \emph{safety} and \emph{adaptivity}. Safety ensures that the proposed estimator never performs worse than a labeled-only baseline, regardless of the quality of the predictions. Specifically, for inference, the SADA estimator guarantees an asymptotic variance no larger than that of the estimator using labeled data only, a property that is generally not ensured by PPI \citep{AngelopoulosBatesFannjiangJordanEtAl2023} or PPI++ \citep{AngelopoulosDuchiZrnic2024}. For prediction, the excess risk of the SADA estimator converges at a rate no slower than that of the labeled-only baseline. 
Adaptivity means that our algorithm automatically aggregates multiple predictions by assigning them appropriate weights, leading to more efficient inference or improved predictive accuracy. 
\label{rev:eff1}

Our contributions are threefold. First, we propose a principled and fully data-driven procedure for aggregating multiple predicted labels of uncertain quality in a semi-supervised learning setting, accompanied by theoretical guarantees, which greatly enhances and broadens the applicability of PPI-based methods in practice. 
Second, we derive the asymptotic representations and excess risk bounds for the SADA estimator, which make explicit how the method operates, demonstrate how it protects against potentially low-quality predictions, and clarify how it automatically borrows strength from informative predictions without requiring prior knowledge or formal testing of the quality of each model. 
Third, we illustrate these properties through intuitive explanations, simulation studies, and two real-data applications with distinct tasks that demonstrate the practical usefulness of the approach. We also provide a user-friendly R package, \texttt{sada}, to facilitate implementation of the proposed algorithm.

\subsection{Related work}

\paragraph*{Semi-supervised learning (SSL)}

Over the past decades, a wide range of SSL algorithms has been proposed. These methods differ in the assumptions, in how they utilize unlabeled data, and in their relationship to supervised learning approaches \citep{vanEngelenHoos2020}. Broadly, these algorithms can be categorized into two types:  \emph{inductive} and \emph{transductive} methods.  Inductive methods, similar to supervised learning, use a pre-trained model to assign labels to unlabeled data. Examples include self-training \citep{Yarowsky1995,Lee2013,BerthelotCarliniGoodfellowPapernotEtAl2019,BerthelotCarliniCubukKurakinEtAl2020}, co-training \citep{BlumMitchell1998,WangZhou2010,DengGuo2011}, pseudo-labeled boosting methods \citep{Zhou2012}, unsupervised preprocessing \citep{SheikhpourSarramGharaghaniChahooki2017}.
In contrast, transductive methods do not produce a generalizable model; instead, they predict labels by directly propagating information through connections between data points.  It typically defines a graph over all data points, both labeled and unlabeled, encoding the pairwise similarity of data points with possibly weighted edges \citep{JebaraWangChang2009,LiuWangChang2012,SubramanyaTalukdar2014}.
Additionally, significant progress has been made in recent years to understand, as well as how to leverage, the statistical benefits of the unlabeled data.
\citet{ChakraborttyCai2018} and \citet{AzrielBrownSklarBerkEtAl2022} studied linear regression problems within the SSL framework and proposed estimators that are more efficient than ordinary least squares (OLS) which relies solely on labeled data. \citet{SongLinZhou2023} further extended this framework to general M-estimation problems. The methodology has also been adapted to high-dimensional settings, where the number of features exceeds the sample size \citep{ZhangBrownCai2019,CaiGuo2020,ZhangBradic2022,DengNingZhaoZhang2024,ying2026dependable}.
In particular, \cite{DengNingZhaoZhang2024} established a minimax lower bound, constructed a modified score based on a conditional mean model, and further developed a safe two-step refitting procedure.\label{rev:ssl1}
Applications of SSL have expanded beyond statistical models to include both 2D computer vision tasks \citep{JeongLeeKimKwak2019,LiuMaHeKuoEtAl2021,ZhouGeLiuMaoEtAl2022} and 3D object detection problems \citep{WangCongLitanyGaoEtAl2021,ParkXuZhouTomizukaEtAl2022,LiuGaoLiuLiEtAl2023}. 

\paragraph*{Prediction-powered inference (PPI)} 
In the past few years, a growing body of research has focused on enhancing statistical inference by incorporating predictions from black-box models \citep{WangMcCormickLeek2020,MotwaniWitten2023}.   
In particular, \citet{AngelopoulosBatesFannjiangJordanEtAl2023} introduced  \emph{prediction-powered inference} (PPI), a framework that enables valid inference even when the predictive model is of low quality. However, PPI might be less efficient than the labeled-only baseline, motivating follow-up works aimed at improving its efficiency or combining it with other statistical and machine learning ideas. Examples include PPI++ \citep{AngelopoulosDuchiZrnic2024}, cross PPI \citep{ZrnicCandes2024}, stratified PPI \citep{FischMaynezHoferDhingraEtAl2024} and recalibrated PPI \citep{JiLeiZrnic2025}.

In related work, \citet{ZhuDingJacobsonWuEtAl2023} proposed a doubly robust self-training method that achieves faster convergence rates when predictions are highly accurate. \citet{MiaoMiaoWuZhaoEtAl2025} introduced a post-prediction adaptive inference approach that ensures valid statistical inference without relying on assumptions about the predictions. \citet{GanLiangZou2024} explored a broader class of imputed loss functions to enhance modeling flexibility and efficiency. \citet{GronsbellGaoShiMcCawEtAl2025} focused on inference under squared error loss, situating PPI within the broader context of semiparametric theory. 
\citet{BartolomeisAbadWangDonhauserEtAl2025} introduced a framework that integrates the predictions from multiple foundation models with randomized experiments while preserving valid statistical inference. 

\paragraph*{Missing data and causal inference}
SSL is also closely related to missing data and causal inference \citep{Rubin1974,Rubin1976}. When missingness is at random,  a rich variety of methods has been conducted to handle this, such as likelihood-based inference \citep{Ibrahim1990}, imputation \citep{Rubin2004}, and semiparametric methods \citep{RobinsRotnitzkyZhao1994,ZhaoLipsitzLew1996}. For more complex missing-not-at-random settings, identification and estimation methods are also well developed, such as outcome selection models \citep{Heckman1979}, graphical models \citep{Fay1986,MaGengHu2003}, sensitivity analysis techniques \citep{RobinsRotnitzkyScharfstein2000}, and approaches based on instrumental variables or shadow variables \citep{WangShaoKim2014,ZhaoShao2015,ZhaoMa2022,MiaoLiuLiTchetgenTchetgenEtAl2024}.

\subsection{Paper organization}
The rest of this paper is organized as follows. 
Section~\ref{sec:setup} introduces the problem setup and data structure.
Section~\ref{sec:sada_inference} presents the SADA approach for inference and develops its theoretical properties, while Section~\ref{sec:sada_prediction} adapts the procedure to the prediction task.
Section~\ref{sec:simulation} constructs simulation studies to evaluate the finite-sample performance of the proposed methods.
Section~\ref{sec:real_data} applies the proposed methods to two real-data analyses with distinct tasks.
Section~\ref{sec:discussion} concludes the paper. 
All proofs are provided in the Supplementary Material.

\subsection{Notations}
The following notations are used throughout the paper.
All vectors are assumed to be column vectors unless otherwise specified.
Let $\u^{\otimes 2}=\u\u\trans$ for a vector $\u$.
We denote $\A\preceq\B$ for two symmetric square matrices $\A$ and $\B$ if $\B-\A$ is positive semi-definite. 
We generally use capital letters to denote random variables, calligraphic letters to denote their support, and lowercase letters to denote their realizations. We use $\mP$ to denote the probability measure and $\mE$ the expectation. 
For two random vectors $\U$ and $\V$, let $\cov(\U,\V)=\mE[\{\U-\mE(\U)\}\{\V-\mE(\V)\}\trans]$ and $\var(\U)=\cov(\U,\U)$. 

\section{Setup}\label{sec:setup}
In this paper, we consider two types of scientific goals: inference and prediction.
Let $Y$ be a scalar outcome and $\X$ be a vector of covariates.
For the inference problem, the target is to estimate and draw inference on a $p$-dimensional parameter, $\btt^*\in\Theta\subset\mR^p$, defined through an estimating equation
\begin{align}\label{eq:EE}
\mE\{\s(\X,Y;\btt^*)\}=\0,
\end{align}
where $\s(\x,y;\btt)$ is a user-specified $p\times 1$ vector-valued function. This formulation is broad and includes many commonly studied parameters, such as outcome means, quantiles, least squares coefficients, and other functionals of $\X$ and $Y$, including the unique minimizer of a loss function or the maximizer of a criterion function \citep{Vaart1998}. The goal is to construct an estimator of $\btt^*$ that is as close as possible to its truth.

By contrast, a prediction problem is defined by a hypothesis family $\{f_\btt:\calX\to\calY\}$ parameterized by $\btt\in\Theta$, together with a loss function $\ell(f_\btt(\x),y)$ that measures the discrepancy between the model's output and the true label. For example, in a binary classification problem, the cross-entropy loss is given by $\ell(f_\btt(\x),y)=-y\log\{f_\btt(\x)\}-(1-y)\log\{1-f_\btt(\x)\}$ with $f_\btt(\x)\in(0,1)$.
Define $\calR(\btt) = \mE\{\ell(f_\btt(\X), Y)\}$ as the expected prediction loss, and $$\btt^* = \amin_{\btt\in\Theta} \calR(\btt),$$ as the optimal prediction parameter.
The goal is to construct an estimator $\wh\btt$ such that the excess risk, $\calR(\wh\btt)-\calR(\btt^*)$, is as small as possible.

We consider a standard SSL setting in which only a subset of the sample is labeled with the ground-truth outcome $Y$. Denote the data as $\calL\cup\calU$, where $\calL=\{(\x_i,y_i),i=1,\dots,n\}$ is labeled and $\calU=\{\x_i,i=n+1,\dots,N\}$ unlabeled. Assume that $\calL$ and $\calU$ are independent and share the same marginal distribution of $\X$. 

\begin{figure}
	\centering
	\includegraphics[width=\textwidth]{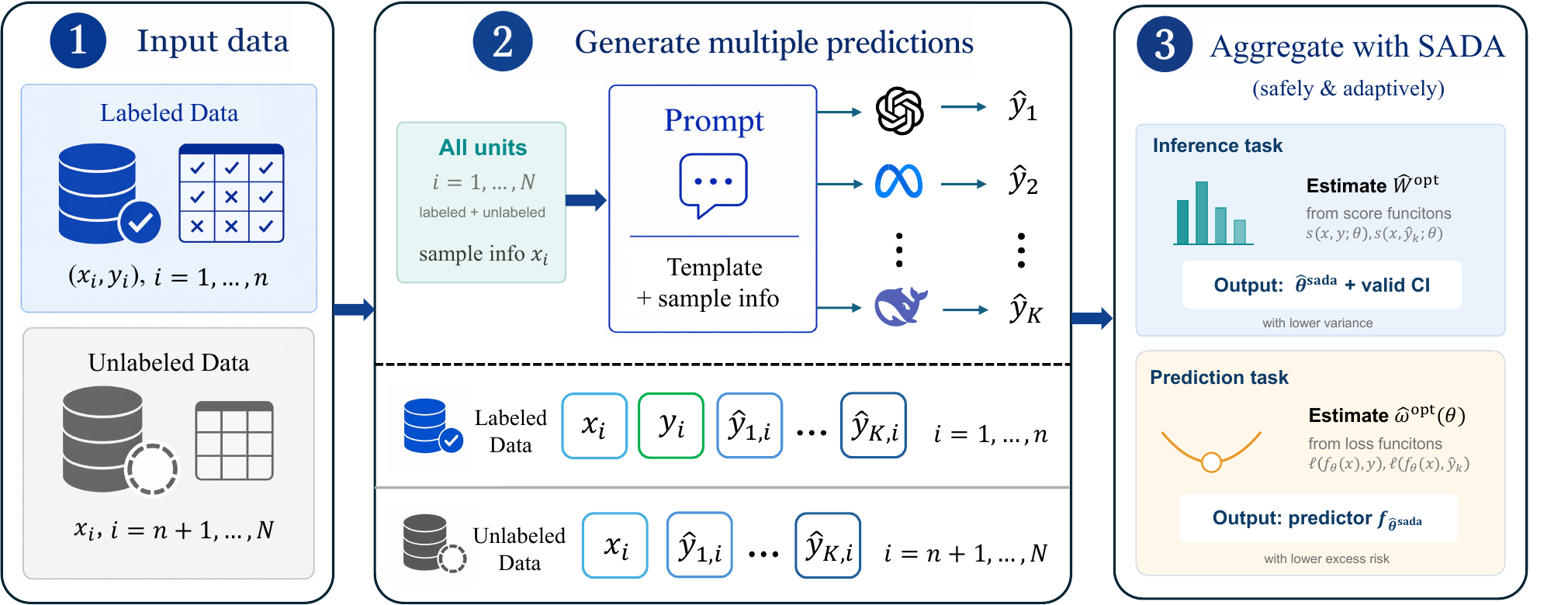} 
    \caption{Protocol for the SADA aggregation procedure.}
	\label{fig:process}
\end{figure}

\label{rev:K_fixed}
In addition, we consider the availability of predicted labels from $K$ black-box models, denoted by $\wh\Y=(\wh Y_1, \ldots, \wh Y_K)\trans$. Throughout, we assume that $K$ is fixed and does not grow with the sample size. These predictions may originate from any low-cost source: crowdsourced annotators, weak-supervision pipelines, measurement-error surrogates, model ensembles, traditional machine learning models, or large pre-trained models whose quality may be uncertain. We do not require the prediction $\wh Y_k$ to share the same magnitude or functional form, either with one another or with the true outcome $Y$. For example, when $Y$ is a binary label, each $\wh Y_k$ may be either categorical or continuous. We allow the generation process of $\wh Y_k$ to be a black box, potentially depending not only on the observed feature $\X$ but also on some latent or unobserved variables. Predicted labels are available for all subjects, both labeled and unlabeled, and are denoted by $\wh \y_i = (\wh y_{1,i}, \ldots, \wh y_{K,i})\trans$, for each subject $i=1,\ldots, N$. We assume that the distribution of $\wh\Y$ is the same for the labeled and unlabeled samples. The data structure and overall generating process are illustrated in \Cref{table:data_structure} and \Cref{fig:process}.

\begin{table}
	\caption{Data structure in semi-supervised learning with multiple sets of predicted labels.}
	\label{table:data_structure}
	\centering
    \begin{tabular}{c c c c l}
        \toprule
        & Unit & Feature & Label & Multiple predicted labels \\
        \midrule
        \multirow{3}{*}{Labeled data $\calL$}  
        & 1 & $\x_1$ & $y_1$ & $\wh\y_1=(\wh y_{1,1},\dots,\wh y_{K,1})\trans$ \\
        & $\vdots$ & $\vdots$ & $\vdots$ & $\quad\vdots$ \\
        & $n$ & $\x_n$ & $y_n$ & $\wh\y_n=(\wh y_{1,n},\dots,\wh y_{K,n})\trans$ \\
        \midrule
        \multirow{3}{*}{Unlabeled data $\calU$}  
        & $n+1$ & $\x_{n+1}$ & ? & $\wh\y_{n+1}=(\wh y_{1,n+1},\dots,\wh y_{K,n+1})\trans$ \\
        & $\vdots$ & $\vdots$ & $\vdots$ & $\quad\vdots$ \\
        & $N$ & $\x_N$ & ? & $\wh\y_N=(\wh y_{1,N},\dots,\wh y_{K,N})\trans$ \\
        \bottomrule
    \end{tabular}%
\end{table}

\section{SADA for inference}\label{sec:sada_inference}
In this section, we consider the estimation and statistical inference of $\btt^*$ defined in \eqref{eq:EE}. Since the outcome is observed only in the labeled set $\calL$, a naive estimator, $\wh\btt^\nv$, based solely on $\calL$, solves
\begin{align*}
  \frac{1}{n}\sum_{i=1}^{n} \s(\x_i, y_i;\btt) =\0.
\end{align*}
Assume that the Hessian matrix $\H=\mE\{\partial \s(\X, Y;\btt^*)/\partial\btt\trans\}$ exists and is nonsingular. Under some regularity conditions (see \Cref{ass:regularity}), $\wh\btt^\nv$ is consistent and asymptotically normal, and its mean squared error, up to a negligible term, is 
\begin{align}\label{eq:Sigmanv}
    \mE\{(\wh\btt^\nv-\btt^*)^{\otimes 2}\} = \frac1n \H^{-1}\bSig_\nv\H^{-\T},
\end{align}
where $\bSig_\nv=\var\{\s(\X,Y;\btt^*)\}$. Based on \eqref{eq:Sigmanv}, valid confidence intervals for $\btt^*$ can be constructed using $\wh\btt^\nv$. However, since the naive estimator uses only the labeled data, it can be substantially less efficient, especially when $n$ is small relative to $N$.

To exploit the unlabeled data, we construct a family of unbiased estimators, $\wh\btt(\calW)$, indexed by the tuning parameter $\calW=(\calW_1\trans,\calW_2\trans,\dots,\calW_K\trans)\trans\in\mR^{(Kp)\times p}$, which solves
\begin{align}\label{eq:generalEE}
  \frac{1}{n}\sum_{i=1}^{n} \s(\x_i, y_i;\btt)+  \sum_{k=1}^K  \calW_k\trans \left\{\frac{1}{N-n}  \sum_{i=n+1}^{N} \s(\x_i, \wh y_{k,i};\btt) - \frac{1}{n} \sum_{i=1}^{n} \s(\x_i, \wh y_{k,i};\btt)\right\} =\0.
\end{align}
Noting that the augmented term has zero expectation regardless of the choice of $\calW$ or the performance of $\wh\y$, therefore, equation \eqref{eq:generalEE} is always a feasible estimating equation for $\btt^*$. In particular, the naive estimator $\wh\btt^\nv$ is a special case with $\calW=\0$. When there is only one prediction $(K=1)$: (i) if $\calW=\I$, $\wh\btt(\calW)$ reduces to the PPI estimator \citep{AngelopoulosBatesFannjiangJordanEtAl2023}; (ii) if $\calW=\omega\I$ with the tuning parameter $\omega$ selected optimally, it reduces to the PPI++ estimator \citep{AngelopoulosDuchiZrnic2024}. However, PPI and PPI++ do not provide a principled way to handle settings with multiple predictions, and even in the special case $K=1$, they do not in general achieve optimality within the family \eqref{eq:generalEE} in terms of the smallest possible asymptotic variance.
For inference purposes, we therefore propose to identify the {optimal} estimator within the family \eqref{eq:generalEE} by minimizing the mean squared error of $\wh\btt(\calW)$. 

\begin{proposition}\label{prop:opt_weight}
  Assume that $\var\{\calS(\X,\wh\Y;\btt^*)\}$ is nonsingular.
    Among the family of estimators $\wh\btt(\calW)$ solving \eqref{eq:generalEE}, the optimal tuning parameter, $\calW^\opt$, that minimizes the mean squared error loss such that $\mE[\{\wh\btt(\calW^\opt)-\btt^*\}^{\otimes 2}]\preceq \mE[\{\wh\btt(\calW)-\btt^*\}^{\otimes 2}]$ for all $\calW\in\mR^{(Kp)\times p}$ is  
\begin{align*}
  \calW^\opt = \frac{N-n}{N} \var\{\calS(\X,\wh\Y;\btt^*)\}^{-1} \mE\{\calS(\X,\wh\Y;\btt^*)\s(\X,Y;\btt^*)\trans\},
\end{align*}
where $\calS(\x,\wh\y;\btt)= \{\s(\x,\wh y_1;\btt)\trans,\cdots,\s(\x,\wh y_K;\btt)\trans\}\trans$.
\end{proposition}

The proof of \Cref{prop:opt_weight} is given in Supplementary Section~\ref{sec:app_pf_prop_opt_weight}.
\Cref{prop:opt_weight} implies that the optimal weight $\calW^\opt$ uniquely exists and $\wh\btt(\calW^\opt)$ consistently attains asymptotic variance no larger than that of the labeled-only estimator, regardless of the quality of predictions.
In practice, the optimal weight, $\calW^\opt$, can be replaced by a consistent estimator, for example, 
\begin{align*}
  \wh\calW^\opt = \frac{N-n}{N}
    \bigg[\frac{1}{N}\sum_{i=1}^N \{\calS(\x_i,\wh\y_i;\wh\btt)-\bar\calS_N(\wh\btt)\}^{\otimes 2}\bigg]^{-1}
  \bigg\{\frac{1}{n}\sum_{i=1}^n \calS(\x_i,\wh\y_i;\wh\btt)\s(\x_i,y_i;\wh\btt)\trans \bigg\},
\end{align*}
where $\bar\calS_N(\btt)=N^{-1}\sum_{i=1}^{N}\calS(\x_i,\wh\y_i;\btt)$, and $\wh\btt$ is a consistent estimator of $\btt^*$, e.g., $\wh\btt^\nv$.  We denote the resulting estimator $\wh\btt^\sada=\wh\btt(\wh\calW^\opt)$ as the SADA estimator. 
Note that, without loss of generality, we assume that $\var\{\calS(\X,\wh\Y;\btt^*)\}$ is nonsingular. 
This condition is imposed only to simplify notation. In degenerate cases where the covariance matrix is singular, for example when some components of the prediction vector are perfectly collinear, the inverse in the optimal weight formula can be simply replaced by a generalized inverse (e.g., the Moore-Penrose inverse). 
The use of a generalized inverse does not affect the final estimator, since the redundant directions induce the same linear combination of the estimating functions and hence yield the same SADA estimator.

Below, we build intuition for how the optimal weight operates and illustrate these properties using a mean-estimation example in \Cref{ssec:mean_estimation}, and formalize them in \Cref{ssec:inference_properties}.

\subsection{Illustration with mean estimation}\label{ssec:mean_estimation}
We elaborate the intuition behind our approach using the example of mean estimation. Consider $\s(\x,y;\theta)=y-\theta$, which corresponds to the estimand $\theta^*= \mE(Y)$, the outcome mean. The naive estimator is simply the sample mean, $\wh\theta^\nv=n^{-1}\sum_{i=1}^{n}y_i$.
To improve efficiency by leveraging $K$ machine learning-predicted outcomes $\wh\y_i$, we consider the family of unbiased estimators $\wh\theta(\bm\omega)$, indexed by weights $\bm\omega=(\omega_1,\dots,\omega_K)\trans\in\mR^K$, obtained by solving \eqref{eq:generalEE}, namely,
\begin{align*}
  \wh\theta(\bm\omega) =  \frac{1}{n} \sum_{i=1}^{n}  y_i   + \sum_{k=1}^K \omega_k \left( \frac{1}{N-n}  \sum_{i=n+1}^{N} \wh y_{k,i} - \frac{1}{n} \sum_{i=1}^{n} \wh y_{k,i}\right) .
\end{align*}
As noted previously, when $\bm\omega=\0$, it reduces to the naive estimator $\wh\theta^\nv$. When there is only one prediction, i.e., $K=1$, and $\omega=1$, it turns out to be the PPI estimator \citep{AngelopoulosBatesFannjiangJordanEtAl2023}: 
\begin{align*}
\wh\theta^\ppi=  \frac{1}{N-n}\sum_{i=n+1}^{N} \wh y_{i} + \frac{1}{n}\sum_{i=1}^{n}  \left( y_i-\wh y_i\right).
\end{align*} 
Assuming that $\var(\wh\Y)$ is positive definite. To find the \emph{best} estimator among the family, we can compute the variance, equivalently, the mean squared error, of $\wh\theta(\bm\omega)$:
\begin{align}\label{eq:var_mean}
 \mE[\{\wh\theta(\bm\omega)-\theta^*\}^2] = \frac{1}{n}\var(Y) + \underbrace{\frac{N}{n(N-n)}\bm\omega\trans \var(\wh\Y) \bm\omega - \frac{2}{n} \bm\omega\trans \cov(\wh\Y,Y)}_{\text{additional variance term}}.
\end{align}
The first term of \eqref{eq:var_mean} is the variance of the naive estimator, while the second captures the variance contributed by leveraging ML predictions. Notice that \eqref{eq:var_mean} is a quadratic form of $\bm\omega$, which achieves the minimum at 
\begin{align}\label{eq:w_opt_mean}
  \bm\omega^\opt = \frac{N-n}{N} \var(\wh\Y)^{-1}\cov(\wh\Y,Y).
\end{align}
The optimal weight $\bm\omega^\opt$ is fully determined by and can be easily estimated from the available data.
In practical applications, one can estimate $\bm\omega^\opt$ as
  \begin{align*}
  \wh{\bm\omega}^\opt = \frac{N-n}{N}  
  \bigg\{\frac{1}{N} \sum_{i=1}^N (\wh\y_i-\overline{\wh\y})(\wh\y_i-\overline{\wh\y})\trans\bigg\}^{-1}
  \bigg\{\frac{1}{n}\sum_{i=1}^n (\wh\y_i-\overline{\wh\y})(y_i-\overline{y})\bigg\},
\end{align*}
where $\overline{y} = n^{-1}\sum_{i=1}^{n} y_i$, and $\overline{\wh\y} = N^{-1}\sum_{i=1}^{N} \wh\y_i$.
Then, the SADA estimator is given by $\wh\theta^\sada=\wh\theta(\wh{\bm\omega}^\opt)$.
In what follows, for certain theoretical analyses and illustrations, we do not distinguish between the estimator $\wh\theta(\bm\omega^\opt)$ and the SADA estimator $\wh\theta^\sada$, as they are asymptotically equivalent. We adopt this simplification when it does not cause confusion. We next demonstrate two key properties of the SADA estimator: \emph{safety} and \emph{adaptivity}.

\paragraph*{Safety} Based on \eqref{eq:var_mean} and \eqref{eq:w_opt_mean}, one can compute that 
\begin{align}\label{eq:var_mean_opt}
  \mE \{(\wh\theta^\sada-\theta^*)^2\}
  = \underbrace{\frac{1}{n}\var(Y)}_{\text{variance of } \wh\theta^\nv} - \underbrace{\left(\frac{1}{n}-\frac{1}{N}\right) \cov(\wh\Y,Y)\trans \var(\wh\Y)^{-1}\cov(\wh\Y,Y)}_{\text{efficiency gain}}.
\end{align}
The efficiency gain in \eqref{eq:var_mean_opt} is always non-negative regardless of the quality of the predictions. It vanishes to zero if and only if $\cov(\wh\Y, Y) = \mathbf{0}$; that means, none of the predictions are correlated with the ground truth $Y$. In this worst-case scenario, the optimal weight is automatically assigned to zero (i.e., $\bm\omega = \mathbf{0}$), reducing to the naive estimator. 
Except for this worst-case scenario, the SADA estimator offers a guaranteed positive efficiency gain over the naive method and enables more informative inference for $\theta^*$.

\paragraph*{Adaptivity} If any one of the predictions is highly accurate, even without knowing in advance which one it is, the weight automatically picks it up and yields an estimator with either a faster convergence rate or an improved estimation efficiency. 
We elaborate this in two scenarios.

First, consider the case where one of the predictions perfectly matches the ground truth, for example, $\wh Y_1 \equiv Y$. As shown in Supplementary Section~\ref{sec:app_supp_mean}, the optimal weight in this case is $\bm\omega^\opt =  (1,0,\dots,0)\trans \cdot (N-n)/N$.
This means that the algorithm selects the most accurate prediction, $\wh Y_1$, for estimation while assigning zero weight to the less informative ones. Consequently, the SADA estimator turns out to be $N^{-1}\sum_{i=1}^{N}\wh y_{1,i}$, and $\mE\{(\wh\theta^\sada-\theta^*)^2\}=N^{-1}\var(Y)$, the same as the oracle estimator that knows the ground truth of the unlabeled data. In this case, the SADA estimator converges at a faster rate of $N^{-1/2}$.

Second, consider the restricted case where the predictions are deterministic functions of the available feature $\X$, i.e., $Y_k=\wh f_k(\X)$. In this setting, the best possible prediction of the outcome is the conditional mean $\mE(Y\mid \X)$, which minimizes the mean squared error between $Y$ and any function $f(\X)$. Assume that $\wh Y_1 \equiv \mE(Y\mid \X)$. As shown in Supplementary Section~\ref{sec:app_supp_mean}, the optimal weight is again $\bm\omega^\opt =  (1,0,\dots,0)\trans \cdot (N-n)/N$.
In this case, $\widehat\theta^{\sada}$ attains the smallest possible asymptotic variance among regular estimators and therefore achieves the semiparametric efficiency bound; see Supplementary Section~\ref{sec:eif} for the definition and Supplementary Section~\ref{sec:app_supp_mean} for its derivation. A formal statement of adaptivity for general estimands is provided in \Cref{thm:adaptivity}.

\begin{remark}[Interpretation]
  We provide an intuitive interpretation of why the proposed algorithm enjoys these properties from the perspective of projection. We can rewrite $\wh\theta^\sada=\wh\theta^\nv+N^{-1}\sum_{i=1}^{N}g(\wh\y_i)-n^{-1}\sum_{i=1}^{n}g(\wh\y_i)$, where $g(\wh\y)=\wh\y\trans\bm\omega^{\opt}=\wh\y\trans\var(\wh\Y)^{-1}\cov(\wh\Y,Y)$ is the $L^2(\mP)$-projection of $Y$ on the space linearly spanned by $\wh\Y$. The projection ensures the \emph{safety} and \emph{adaptivity} of the SADA estimator. Specifically, 
  (i) when all predictions $\wh\Y$ are inaccurate and uncorrelated with the ground truth $Y$, the projection $g(\wh\y)$ shrinks to zero, reducing the SADA estimator to the naive one--but never performing worse than it;
  and (ii) when some prediction, for example $\wh Y_1$, most closely fits the ground truth $Y$, the projection of $Y$ to $\wh\Y$ turns out to be $\wh Y_1$, resulting in an estimator $\wh\theta^\sada=\wh\theta^\nv+N^{-1}\sum_{i=1}^{N}\wh y_{1,i}-n^{-1}\sum_{i=1}^{n}\wh y_{1,i}$. 
  In addition, if $\wh Y_1\equiv Y$ then $\wh\theta^\sada=N^{-1}\sum_{i=1}^{N}\wh y_{1,i}$ achieves a faster convergence rate; if $\wh Y_1\equiv \mE(Y\mid \X)$ then $\wh\theta^\sada$ becomes the semiparametrically efficient estimator of $\mE(Y)$.
\end{remark}

\begin{remark}[Comparison with PPI++]
  In case of the mean estimation (or, more generally, a scalar parameter $\theta^*$) with $K=1$, the SADA estimator $\wh\theta^\sada$ is conceptually equivalent to the PPI++ estimator \citep{AngelopoulosDuchiZrnic2024}. However, for a vector-valued parameter, the PPI++ estimator does not in general guarantee a smaller variance than the labeled-only estimator and is thus less efficient than SADA. This phenomenon is evident in our empirical results in \Cref{sec:real_data}. In addition, the PPI++ approach cannot simultaneously leverage multiple predictions when $K > 1$.
\end{remark}

\subsection{Theoretical guarantees}\label{ssec:inference_properties}
We now formally establish the theoretical properties of $\wh\btt^\sada$.
We impose the following regularity conditions that ensure identification, smoothness, and well-behaved large-sample properties of the estimator.
\begin{assumption}\label{ass:regularity}
(i) The parameter $\btt^*$ lies in the interior of $\Theta$, and $\Theta$ is compact in $\mR^p$; 
(ii) $\mE\{\s(\X,Y;\btt)\}\neq 0$ for any $\btt\neq\btt^*$; 
(iii) There exists $\delta>0$ such that $\mE\{\sup_{\|\btt-\btt^*\|<\delta}\| \s(\X, Y;\btt)\|^2\}<\infty$ and $\mE\{\sup_{\|\btt-\btt^*\|<\delta}\| \s(\X, \wh Y_k;\btt)\|^2\}<\infty$ for $k=1,\dots,K$; 
(iv) $\s(\x,y;\btt)$ is differentiable with respect to $\btt$ in a neighborhood of $\btt^*$, and 
there exists $\delta>0$ such that
 $\mE\{\sup_{\|\btt-\btt^*\|<\delta}\|\partial \s(\X, Y;\btt)\trans/\partial\btt\|^2\}<\infty$ and $\mE\{\sup_{\|\btt-\btt^*\|<\delta} \allowbreak\|\partial \s(\X, \wh Y_k;\btt)\trans/\partial\btt\|^2\}<\infty$ for $k=1,\dots,K$.
\end{assumption}

The following theorem formalizes the safety property of the SADA estimator and is proved in Supplementary Section~\ref{sec:app_pf_thm_eff_gain}.

\begin{theorem}[Safety] \label{thm:general_eff_gain}
  Under \Cref{ass:regularity}, the SADA estimator $\wh\btt^\sada$ has the asymptotic representation $\sqrt{n} (\wh\btt^\sada - \btt^*)\convd \calN\left( \0,\H^{-1}\bSig_\opt\H^{-\T}\right)$.
  More specifically, for its mean squared error, up to a negligible term, we have
  \begin{align*}
      \mE\{(\wh\btt^\sada - \btt^*)^2\} = \frac1n\H^{-1}\bSig_\opt\H^{-\T},
  \end{align*}
  where $\bSig_\opt=\bSig_\nv - (N-n)/N \cdot \bSig_g$, and
  \begin{align*}
      \bSig_g =  \mE\{\s(\X,Y;\btt^*)\calS(\X,\wh\Y;\btt^*)\trans\} \var\{\calS(\X,\wh\Y;\btt^*)\}^{-1}
 \mE\{\calS(\X,\wh\Y;\btt^*)\s(\X,Y;\btt^*)\trans\}
  \end{align*}
 is always positive semi-definite.
 It implies that $\var(\wh\btt^\sada)\preceq \var(\wh\btt^\nv)$.
\end{theorem}
\Cref{thm:general_eff_gain} shows that the SADA estimator enjoys a guaranteed efficiency gain over the naive estimator, regardless of the quality of $\wh\Y$. This ensures valid and more informative inference for $\btt^*$.
The following theorem formalizes the adaptivity property and is proved in Supplementary Section~\ref{sec:app_pf_thm_adap}.

\begin{theorem}[Adaptivity]\label{thm:adaptivity}
  (i) Suppose $\wh Y_k\equiv Y$ for some $k$ and \Cref{ass:regularity} holds, then we have $\bSig_g=\bSig_\nv$, and
the SADA estimator $\wh\btt^\sada$ has the asymptotic representation $\sqrt{N} (\wh\btt^\sada - \btt^*)\convd \calN\left( \0,\H^{-1}\bSig_\nv\H^{-\T}\right)$.
  More specifically, for its mean squared error, up to a negligible term, we have
  \begin{align*}
      \mE\{(\wh\btt^\sada - \btt^*)^2\} = \frac1N\H^{-1}\bSig_\nv\H^{-\T}.
  \end{align*}
  Note that this is the same as the oracle estimator which knows the ground truth of the unlabeled data;
  (ii) Suppose $n/N\to \pi\in(0,1)$ as $n,N\to\infty$. Assume $\wh Y_k=\wh f_k(\X)$ for $k=1,\dots,K$. Then the EIF for estimating $\btt^*$, based on data $\calL \cup \calU$, is
  \begin{align*}
  	\bm\Phi_\eif(r,\x,y) = -\H^{-1}\left[r \pi^{-1}\{\s(\x,y;\btt^*)-\bm\mu(\x)\} + \bm\mu(\x)\right],
  \end{align*}
  where $\bm\mu(\x)=\mE\{\s(\x,Y;\btt^*)\mid\x\}$. Suppose $\s(\x,\wh y_{k'};\btt^*) =\s(\x,\wh f_{k'}(\x);\btt^*)\equiv \bm\mu(\x)$ for some $k'$, then we have $\sqrt{N}(\wh\btt^\sada - \btt^*)\convd \calN\{\0,\mE(\bm\Phi_\eif\bm\Phi_\eif\trans)\}$,
  which attains the semiparametric efficiency bound.
\end{theorem}

\Cref{thm:adaptivity}(i) shows that, in the oracle scenario where one of the predictions coincides with the outcome ground truth, SADA increases the effective sample size from $n$ to $N$ and thereby attains a faster convergence rate.
\Cref{thm:adaptivity}(ii) shows, when all predictions are functions of the covariates $\X$ and one score function equals the corresponding conditional mean, the estimator attains the semiparametric efficiency bound. Importantly, we never need to know in advance which prediction is the best; the entire procedure is data-driven, which makes it appealing in practice.

\section{SADA for prediction}\label{sec:sada_prediction}
The previous section focuses on estimation and inference for the parameter $\btt^*$, where the primary goal is to obtain an unbiased and more efficient estimator. In other applications, such as image classification, practitioners are primarily concerned with predictive accuracy rather than parameter estimation itself, and in such cases the SADA framework needs to be adapted accordingly.
Intuitively, even if one estimator, say, $\wh\btt_1$, is closer to $\btt^*$ than the other estimator $\wh\btt_2$ in mean squared error, it does not in general imply a smaller predictive risk; that is, $\calR(\wh\btt_1)\le \calR(\wh\btt_2)$. Motivated by this distinction, in this section we slightly modify the SADA procedure to explicitly target minimization of the predictive risk $\calR(\btt)$ and discuss how it can be designed to achieve better prediction performance.

To build intuition, consider an unbiased estimator $\wh\calR(\btt)$ of $\calR(\btt)$ such that $\mE\{\wh\calR(\btt)\}=\calR(\btt)$. Define $\wh\btt=\amin_{\btt\in\Theta} \wh\calR(\btt)$. Then, in a heuristic sense, the excess risk $\calR(\wh\btt)-\calR(\btt^*)  \le \calR(\wh\btt) - \wh\calR(\wh\btt) + \wh\calR(\btt^*) -\calR(\btt^*) \le 2\sup_{\btt\in\Theta}|\wh\calR(\btt)-\calR(\btt)|$, where the first inequality follows from $\wh\calR(\wh\btt)\le \wh\calR(\btt^*)$ by definition. Under some regularity conditions, standard uniform concentration results indicate that this supremum depends on the complexity of $\Theta$, the variance scale of $\wh\calR$, and the tail behavior of the empirical loss. Accordingly, once class complexity is fixed, reducing the variance of $\wh\calR$ is a direct and effective way to tighten the uniform deviation and lower the excess risk.

Let $\wh\calR^{\nv}(\btt)=n^{-1}\sum_{i=1}^{n} \ell(f_\btt(\x_i),y_i)$ denote the naive empirical loss function which only uses the labeled data, and $\wh\btt^\nv = \amin_{\btt\in\Theta} \wh\calR^{\nv}(\btt)$. We consider a set of empirical loss functions, indexed by $\bm\omega = (w_1, w_2,\dots,w_K)\trans$,
\begin{align}\label{eq:aug_loss}
\wh\calR(\btt,\bm\omega) = \frac{1}{n}\sum_{i=1}^{n} \ell(f_\btt(\x_i),y_i) +  \bm\omega \trans \left\{\frac{1}{N-n}  \sum_{i=n+1}^{N} \calL(f_\btt(\x_i), \wh \y_{i}) - \frac{1}{n} \sum_{i=1}^{n} \calL(f_\btt(\x_i), \wh \y_{i})\right\},
\end{align}
where $\calL(f_\btt(\x), \wh \y) = (\ell(f_\btt(\x),\wh y_{1}),\ell(f_\btt(\x),\wh y_{2}),\dots,\ell(f_\btt(\x),\wh y_{K}) )\trans$. Noting that the second term has zero expectation for any $\bm\omega$ and any distribution of $\wh\y$, therefore, equation \eqref{eq:aug_loss} always defines an unbiased estimator for $\calR(\btt)$. As a special case, when $\bm\omega=\0$, $\wh\calR(\btt,\bm\omega)$ reduces to $\wh\calR^{\nv}(\btt)$. For a fixed $\btt$, we propose to select the optimal loss function from the set by minimizing the mean squared error between $\wh\calR(\btt,\bm\omega)$ and the expected loss $\calR(\btt)$. 
We have 
\begin{align*}
\bm\omega^\opt(\btt) =&~  \amin_{\bm\omega} \mE\{\wh\calR(\btt,\bm\omega)-\calR(\btt)\}^2 \\ 
=&~ \frac{N-n}{N} \var\{\calL(f_\btt(\X), \wh \Y)\}^{-1} \cov\{\calL(f_\btt(\X), \wh \Y), \ell(f_\btt(\X),Y)\}.
\end{align*}
Then, we define $\wh\calR^{\sada}(\btt) = \wh\calR(\btt,\wh{\bm\omega}^\opt(\btt))$, where $\wh{\bm\omega}^\opt$ is a consistent estimator of $\bm\omega^\opt$, and obtain the SADA estimator as $\wh\btt^\sada = \amin_{\btt\in\Theta} \wh\calR^{\sada}(\btt)$. It is worth noting that the optimal weight depends on $\btt$ and, in principle, should be optimized jointly with the parameter within the loss function, which may lead to numerical difficulties and instability in practice. Therefore, for implementation, we recommend an iterative procedure that alternately updates the weight $\bm\omega^\opt$ and the target parameter $\btt^*$, as outlined in \Cref{alg:sada_prediction}.

\begin{algorithm}[htbp]
\caption{SADA for prediction}
\label{alg:sada_prediction}
\begin{algorithmic}[1]
\STATE \textbf{Input:} Observations $(\x_i,y_i,\wh\y_i)_{i=1}^n\cup(\x_i,\wh\y_i)_{i=n+1}^N$, prediction function $f_\btt$, loss function $\ell(f_\btt(\x),y)$, number of iterations $J$.
\STATE Compute the naive estimator $\wh\btt^\nv = \amin_{\btt}  n^{-1}\sum_{i=1}^{n} \ell(f_\btt(\x_i),y_i)$.
\STATE Initialize $\wh\btt^{(0)}=\wh\btt^\nv$.
\FOR{$j = 1, \ldots, J$}
    \STATE Compute the optimal weight 
     \begin{align*}
     \wh{\bm\omega}^{(j)}=\frac{N-n}{N} &\Bigg\{\frac{1}{N}\sum_{i=1}^N \wt\calL_N\left(f_{\wh\btt^{(j-1)}}(\x_i),\wh\y_i\right)\wt\calL_N\left(f_{\wh\btt^{(j-1)}}(\x_i),\wh\y_i\right) \trans\Bigg\}^{-1} \\
      &~~\times\Bigg\{\frac{1}{n}\sum_{i=1}^n  \calL\left(f_{\wh\btt^{(j-1)}}(\x_i),\wh\y_i\right)  \ell\left(f_{\wh\btt^{(j-1)}}(\x_i),y_i\right) \Bigg\},
     \end{align*}
     where $\wt\calL_N\left(f_\btt(\x),\wh\y\right)=\calL\left(f_\btt(\x),\wh\y\right)-N^{-1}\sum_{i=1}^N\calL\left(f_\btt(\x_i),\wh\y_i\right)$.
    \STATE Update the parameter estimate
    \begin{align*}
    \wh\btt^{(j)}=\amin_{\btt\in\Theta}  \frac{1}{n}\sum_{i=1}^{n} \ell(f_\btt(\x_i),y_i) +   \Bigg\{\frac{1}{N-n}  \sum_{i=n+1}^{N} \calL(f_\btt(\x_i), \wh \y_{i}) - \frac{1}{n} \sum_{i=1}^{n} \calL(f_\btt(\x_i), \wh \y_{i})\Bigg\}\trans\wh{\bm\omega}^{(j)}.
    \end{align*}
\ENDFOR
\STATE \textbf{Output:} SADA estimator $\wh\btt^\sada=\wh\btt^{(J)}$ and the corresponding predictor $f_{\wh\btt^\sada}(\x)$.
\end{algorithmic}
\end{algorithm}

\subsection{Theoretical properties}
To investigate the theoretical properties of $\wh\btt^\sada$ in terms of its excess risk, $\calR(\wh\btt^\sada)-\calR(\btt^*)$, we first introduce some notations. For a metric space $(\Theta, d)$, let $N(\epsilon, \Theta,d)$ denote the covering number, i.e., the minimum number of balls of radius  $\epsilon$  with respect to the metric $d$ that cover the space $\Theta$.
For a positive constant $C$, define a class of sub-Gaussian random variables
\begin{align*}
    \calF_C = \left\{ \xi: \text{$\xi$ is $\sigma$-sub-Gaussian for some } \sigma \leq C\sqrt{\var(\xi)}  \right\} .
\end{align*}
We use some simplified notations when no confusion arises. Let $\ell_\btt=\ell(f_\btt(\X),\Y)$, $\wt\ell_\btt=\ell_\btt-\calL(f_\btt(\X), \wh \Y)\trans\bm\omega^\opt(\btt)$, and $\wh\ell_\btt = N/(N-n)(\ell_\btt-\wt\ell_\btt)$.
We endow the parameter space $\Theta$ with three metrics: $d_1(\btt,\btt') =  \sd(\ell_\btt -\ell_{\btt'})$, $d_2(\btt,\btt')=\sd(\wh \ell_\btt-\wh\ell_{\btt'})$ and $d_3(\btt,\btt') = \sd(\wt{\ell}_\btt -\wt{\ell}_{\btt'})$, where $\sd(U) = \sqrt{\var(U)}$ denotes the standard deviation for a random variable $U$.
We impose the following regularity conditions, which control the smoothness and tail behavior of the loss functions and the complexity of the parameter space.

\begin{assumption}\label{ass:complexity}
    (i) The loss function $\ell(f_\btt(\x),y)$ is continuous in $\btt$ and uniformly bounded for all $\btt\in\Theta$ and $(x,y)\in\calX\times\calY$; (ii) There exists a positive constant $C>0$ such that, $\forall \btt, \btt' \in \Theta$, $\ell_\btt-\ell_{\btt'}$, $\wh \ell_\btt-\wh \ell_{\btt'}$ and $\wt{\ell}_\btt -\wt{\ell}_{\btt'} $ all belong to $\calF_C$. (iii) The Dudley's entropy integral is finite for $d_1,d_2$ and $d_3$; that is, for $j=1,2,3,$
    \begin{align*}
        \int_0^\infty \sqrt{\log N(\epsilon, \Theta,d_j)} d \epsilon < \infty.
    \end{align*}
\end{assumption}

\begin{theorem}[Excess risk]\label{thm:excess_risk}
Under \Cref{ass:complexity}, we have that with probability at least $1-\delta$,
\begin{align*}
\calR(\wh\btt^\sada) - \calR(\btt^*)\leq
  \mathcal{O} \left( \sqrt{\frac{1}{n} \log \frac{1}{\delta}}  
    + \sqrt{\frac{1}{n}}  
    + \sqrt{\frac{1}{N} \log \frac{1}{\delta}}
  + \sqrt{\frac{1}{N}} \right).
\end{align*}
In addition, if $\wh Y_k=Y$ for some $k\in\{1,\dots,K\}$, we have 
\begin{align*}
\calR(\wh\btt^\sada) - \calR(\btt^*)\leq \mathcal{O} \left( \sqrt{\frac{1}{N} \log \frac{1}{\delta}} + \sqrt{\frac{1}{N}} \right).
\end{align*}
\end{theorem}

A more precise non-asymptotic bound for $\calR(\wh\btt^\sada)-\calR(\btt^*)$ is given in Theorem~\ref{thm:excess_risk_supp} in Supplementary Section~\ref{sec:supp_pf_thm_prediction}. 
Moreover, Theorem~\ref{thm:naive_minimax_supp} shows that the 
naive estimator attains the minimax rate $n^{-1/2}$ for labeled-only 
prediction. Hence, the first bound in Theorem~\ref{thm:excess_risk} guarantees 
that SADA is no worse than this minimax-rate-optimal labeled-only baseline, 
regardless of the quality of the predictions. 
In contrast, if one of the predictions perfectly fits the ground truth, although it is unknown which one, the SADA estimator achieves an even faster convergence rate, increasing the effective sample size from $n$ to $N$. 

\section{Simulation studies}\label{sec:simulation}
In this section, we conduct simulation studies to evaluate the performance of the proposed method. We consider two data-generating processes. In the first, we study mean estimation to illustrate how SADA adaptively leverages predictions of varying quality. In the second, we consider a regression setting and compare SADA with competing methods across multiple metrics.

\subsection{Mean estimation}\label{ssec:simu_mean}
We generate $Y\sim \calN(\theta^*,1)$ with $\theta^*=0.5$, and generate two predictions $\wh Y_1=\gamma Y+(1-\gamma)\epsilon_1$ and $\wh Y_2=(1-\gamma)Y+\gamma\epsilon_2$, where $\gamma$ is a tuning parameter and $\epsilon_1,\epsilon_2$ are independent white noises drawn from the standard normal distribution. The goal is to estimate and to conduct inference for $\theta^*$.
As $\gamma$ increases from 0 to 1, the quality of the prediction $\wh Y_1$ improves in approximating the ground truth $Y$, while the prediction $\wh Y_2$ becomes less accurate. We generate a dataset of size $N=200$, of which $n=60$ observations are labeled, and repeat the experiment over 1000 Monte Carlo replications.

\begin{figure}[]
    \centering
    \begin{subfigure}{.32\linewidth}
        \centering
        \includegraphics[width=\textwidth]{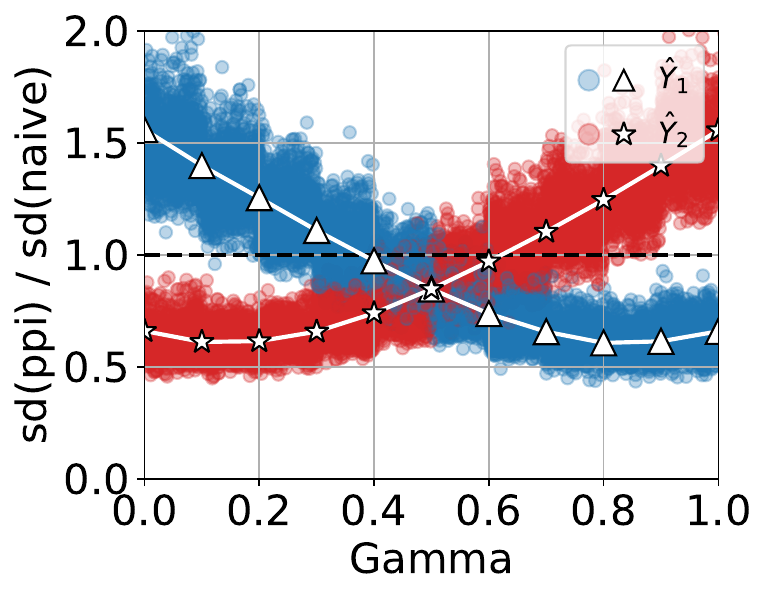}
        \caption{PPI}
        \label{fig:sd_ratio_ppi}
    \end{subfigure}
    \begin{subfigure}{.32\linewidth}
        \centering
        \includegraphics[width=\textwidth]{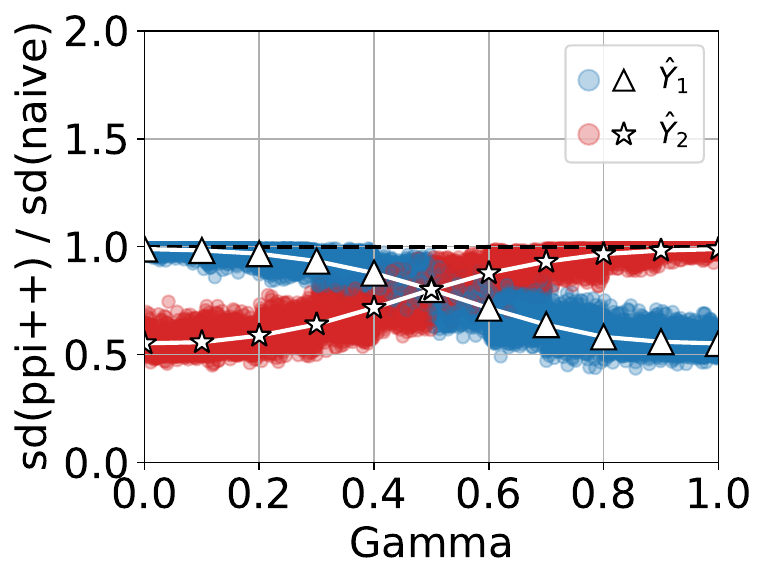}
        \caption{PPI++}
        \label{fig:sd_ratio_ppi++}
    \end{subfigure}
    \begin{subfigure}{.32\linewidth}
        \centering
        \includegraphics[width=\textwidth]{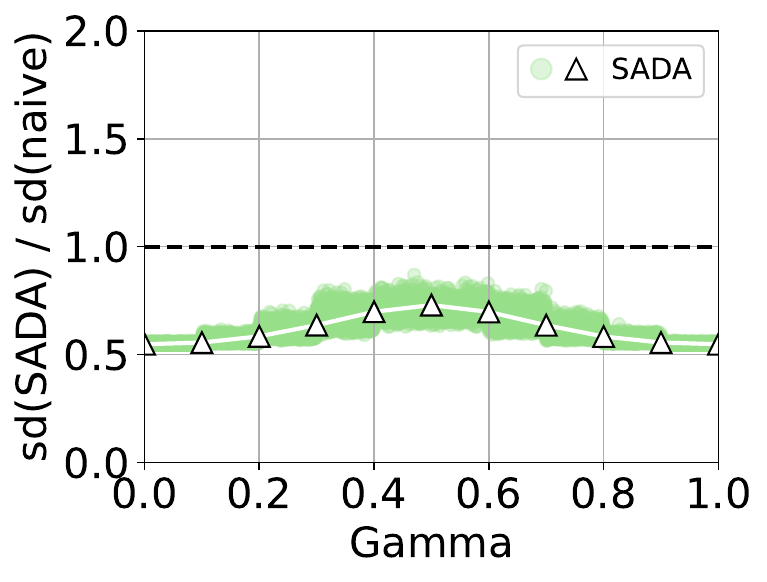}
        \caption{SADA (ours)}
        \label{fig:sd_ratio_sada}
    \end{subfigure}
    \caption{Relative efficiency of different methods compared to the naive method as prediction quality varies. Stars and triangles on the line indicate standard deviations over 1000 replications; scatter points represent those from individual replications.}
    \label{fig:sd_ratio}
\end{figure}

We include naive, PPI, and PPI++ methods for comparison. \Cref{tab:sd_ratio} reports the relative efficiency of each method relative to the naive method, defined as the ratio of their standard deviations across replications as $\gamma$ varies from 0 to 1, and \Cref{fig:sd_ratio} presents the corresponding plots. The results on bias and coverage probability are reported in Supplementary Section~\ref{sec_supp:add_simu}, as they show negligible differences across methods and are not the primary criteria for assessing their performance.
It is shown that, the performance of the PPI estimator is highly sensitive to prediction accuracy: it can perform even worse than the naive estimator when the prediction is poor, as illustrated in \Cref{fig:sd_ratio}(a), where the ratios of standard deviations exceed 1 when $\gamma$ approaches 0 for PPI using $\wh Y_1$ and 1 for PPI using $\wh Y_2$. 
By contrast, PPI++ safeguards against low quality predictions, achieving a variance that is never larger than that of the naive estimator, as shown in \Cref{fig:sd_ratio}(b). However, its efficiency gains arise only when the predictions are informative, and it reduces to the naive estimator when the predictions are noninformative.

In comparison, the SADA estimator exhibits several favorable properties. First, it consistently attains a smaller standard deviation than the naive estimator across the entire range of prediction quality. Second, it adaptively exploits the relative strengths of $\wh Y_1$ and $\wh Y_2$ in the regions where each is more informative, which leads to the most stable performance among all methods considered. Third, it is able to incorporate information from both predictions through data-driven weights. This is particularly evident at $\gamma=0.5$, where SADA has substantially lower standard deviations than PPI++ under either prediction.

\begin{table}
\caption{Relative efficiency (ratio of standard deviations for each method to that of the naive method) under varying prediction quality $\gamma$.}
\label{tab:sd_ratio}
\setlength{\tabcolsep}{3.1pt}
\resizebox{\textwidth}{!}{
\begin{tabular}{@{}ccccccccccccc@{}}
\toprule
Method &  &\multicolumn{10}{c}{Relative efficiency}\\\midrule
   & & $\gamma=0$  & $\gamma=0.1$ & $\gamma=0.2$ & $\gamma=0.3$ & $\gamma=0.4$ & $\gamma=0.5$ & $\gamma=0.6$ & $\gamma=0.7$ & $\gamma=0.8$ & $\gamma=0.9$ & $\gamma=1$   \\[5pt]  
\multirow{2}{*}{PPI} & $\wh{Y}_1$ & 1.549 & 1.396 & 1.247 & 1.104 & 0.970 & 0.842 & 0.736 & 0.653 & 0.608 & 0.610 & 0.654 \\ 
 & $\wh{Y}_2$ & 0.655 & 0.609 & 0.611 & 0.656 & 0.737 & 0.845 & 0.968 & 1.103 & 1.245 & 1.392 & 1.551 \\ 
\multirow{2}{*}{PPI++} & $\wh{Y}_1$ & 0.989 & 0.982 & 0.965 & 0.933 & 0.877 & 0.799 & 0.713 & 0.636 & 0.582 & 0.556 & 0.549 \\ 
 & $\wh{Y}_2$ & 0.549 & 0.556 & 0.585 & 0.639 & 0.715 & 0.801 & 0.876 & 0.932 & 0.965 & 0.982 & 0.989 \\ 
SADA & both & 0.548 & 0.555 & 0.584 & 0.636 & 0.697 & 0.727 & 0.697 & 0.635 & 0.583 & 0.555 & 0.548 \\ 
\bottomrule
\end{tabular}
}
\end{table}

\subsection{Least squares regression}\label{ssec:simu_ols}
We generate $\X=(1,X_1,X_2)\trans$, where $X_j\sim \calN(0,1)$, and $Y = 0.5 + 0.5X_1 + 0.5X_2 + Z_1 + Z_2 + \varepsilon$, where $Z_1 = X_1^2 - 1$, $Z_2 = X_2^2 - 1$, and $\varepsilon \sim \calN(0,1)$. The two terms $Z_1$ and $Z_2$ represent nonlinear components in the conditional mean of $Y$. The parameter of interest is the least squares coefficient $\btt^*=\amin_\btt\mE\{(Y-\X\trans\btt)^2\}$, whose true value is $\btt^* = (0.5,\,0.5,\,0.5)\trans$. To mimic the practical setting in which multiple imperfect predictive models are available, we train two learners on an external independent sample generated from the same data-generating process, and then use the covariates in the analysis sample as inputs to obtain two sets of predictions. Specifically, $\wh Y_1$ is obtained by regressing $Y$ on $(1,X_1,X_2,X_1^2)$, and $\wh Y_2$ is obtained by regressing $Y$ on $(1,X_1,X_2,X_2^2)$. We generate a dataset of size $N=600$, among which $n=180$ observations are labeled, and repeat the experiment over 1000 Monte Carlo replications.

Table~\ref{tab:simu_reg} reports the bias, standard error, coverage probability, and power for each component of $\btt^*$. Here, power is defined as the proportion of replications in which the corresponding coefficient is declared significant at the 5\% level, equivalently, the proportion of replications in which the $95\%$ confidence interval excludes zero. Overall, all methods exhibit small bias across the three coefficients, while the main difference lies in efficiency. The naive estimator has the largest standard errors for all three coefficients.
PPI and PPI++ exploit information from a single auxiliary prediction and therefore reduce the standard errors relative to the naive estimator. However, the improvement is not uniform across all metrics: the average confidence interval length and power are not always better than those of the naive estimator. By contrast, SADA leverages both auxiliary predictions simultaneously and performs best overall. In Table~\ref{tab:simu_reg}, it achieves the smallest standard errors for all three coefficients, and correspondingly has the shortest confidence intervals and the highest power in nearly all cases, while maintaining small bias and coverage close to the nominal level.

\begin{table}[!h]
\centering
\caption{Simulation results for the regression setting. Reported are bias, standard error (SE), coverage probability (CP), average confidence interval length (Len), and power.}
\label{tab:simu_reg}
\setlength{\tabcolsep}{9pt}        
\begin{tabular}{cccccccc}
\toprule
Coefficient & Method &  & Bias & SE & CP & Len & Power \\
\midrule
\multirow{6}{*}{Intercept}
& Naive  & -          & -0.021 & 0.166 & 0.938 & 0.649 & 0.852 \\
& PPI    & $\hat Y_1$ & -0.016 & 0.150 & 0.938 & 0.580 & 0.912 \\
& PPI    & $\hat Y_2$ & -0.014 & 0.143 & 0.954 & 0.580 & 0.938 \\
& PPI++  & $\hat Y_1$ & -0.036 & 0.145 & 0.922 & 0.557 & 0.908 \\
& PPI++  & $\hat Y_2$ & -0.037 & 0.141 & 0.940 & 0.557 & 0.916 \\
& SADA   & both       & -0.023 & 0.110 & 0.936 & 0.436 & 0.996 \\
\midrule
\multirow{6}{*}{$X_1$}
& Naive  & -          & 0.008  & 0.261 & 0.779 & 0.650 & 0.753 \\
& PPI    & $\hat Y_1$ & -0.001 & 0.204 & 0.949 & 0.789 & 0.701 \\
& PPI    & $\hat Y_2$ & 0.008  & 0.251 & 0.950 & 0.982 & 0.542 \\
& PPI++  & $\hat Y_1$ & 0.004  & 0.190 & 0.938 & 0.699 & 0.798 \\
& PPI++  & $\hat Y_2$ & 0.008  & 0.251 & 0.941 & 0.948 & 0.567 \\
& SADA   & both       & 0.003  & 0.164 & 0.923 & 0.599 & 0.879 \\
\midrule
\multirow{6}{*}{$X_2$}
& Naive  & -          & 0.006  & 0.273 & 0.772 & 0.650 & 0.743 \\
& PPI    & $\hat Y_1$ & 0.004  & 0.264 & 0.950 & 0.982 & 0.537 \\
& PPI    & $\hat Y_2$ & 0.002  & 0.207 & 0.937 & 0.786 & 0.698 \\
& PPI++  & $\hat Y_1$ & 0.003  & 0.261 & 0.934 & 0.946 & 0.565 \\
& PPI++  & $\hat Y_2$ & 0.004  & 0.193 & 0.921 & 0.696 & 0.801 \\
& SADA   & both       & 0.002  & 0.165 & 0.912 & 0.596 & 0.886 \\
\bottomrule
\end{tabular}
\end{table}

\section{Real data applications}\label{sec:real_data}
In this section, we apply the proposed methods to two real-data analyses with distinct tasks.
The first is a regression setting based on online requests from Stack Exchange and Wikipedia \citep{Danescu-Niculescu-MizilSudhofJurafskyLeskovecEtAl2013}, where we conduct estimation and inference on the regression coefficient linking perceived politeness to specific linguistic features.
The second is an image classification task based on the ImageNet dataset \citep{russakovsky2015imagenet}.

\subsection{Politeness of online requests}\label{ssec:exp_politeness}
We study how certain linguistic features affect perceived politeness using the dataset from \cite{Danescu-Niculescu-MizilSudhofJurafskyLeskovecEtAl2013}.
The data consist of over ten thousand two-sentence requests collected from Wikipedia and Stack Exchange. Each request was independently annotated by five U.S.-based Amazon Mechanical Turk workers on a 1-25 scale, and the final politeness score is the average of these five ratings. The dataset also includes automatically extracted politeness markers such as gratitude, deference, greetings, sentiment cues, modality choices, and pronominal patterns, obtained by combining dependency parsing with curated lexical resources. In this experiment, we focus on the relationship between politeness and indicative modal features. We select 1000 requests from the dataset and treat the regression coefficient from regressing the politeness score on the indicative modal features as the parameter of interest. Among these requests, we randomly designate 300 as unlabeled data and vary the number of labeled samples $n$ from 50 to 700.

To evaluate the proposed methods, we also obtain predictions for each request from three large language models, GPT-4o, Llama-3-8B, and DeepSeek-V3, by prompting each model to rate the politeness of the given text on a scale of 1-25. For each request, we supply the model with the text under review, relevant background information, and a set of ten independent prompts designed to reduce randomness in the generated scores. The final prediction is computed as the average of the ten outputs. Supplementary Section~\ref{sec:prompt} provides a step-by-step description of the procedure used to generate these predictions.

\begin{figure}
    \centering
    \includegraphics[width=1.0\textwidth]{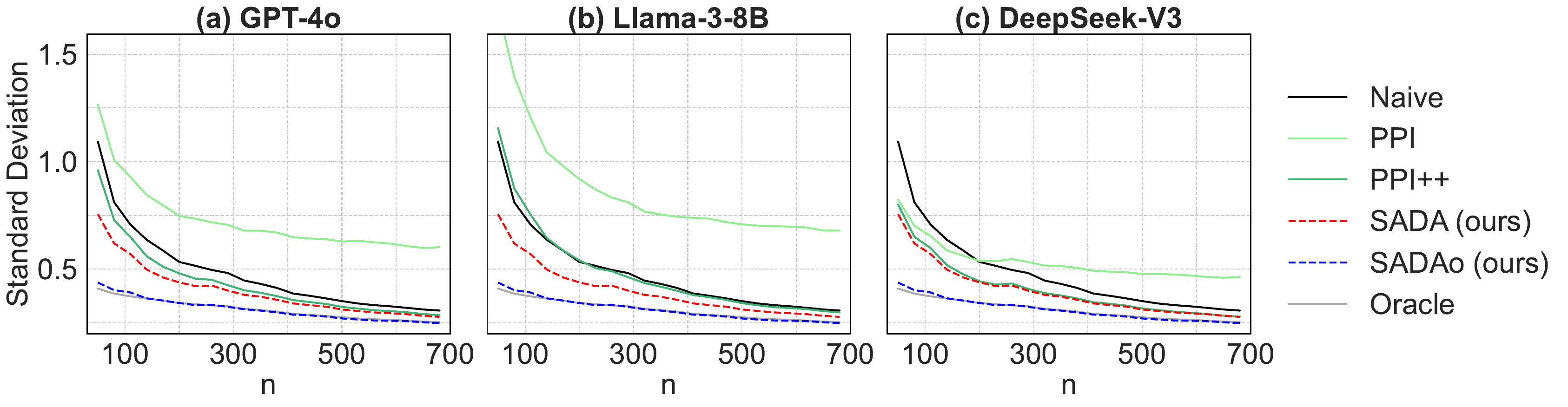}
    \caption{Comparison of standard deviations of different methods leveraging various prediction strategies. The estimand is the regression coefficient of politeness score on indicative modal features.}
    \label{fig:politeness}
\end{figure}

\begin{figure}
    \centering
    \includegraphics[width=1.0\textwidth]{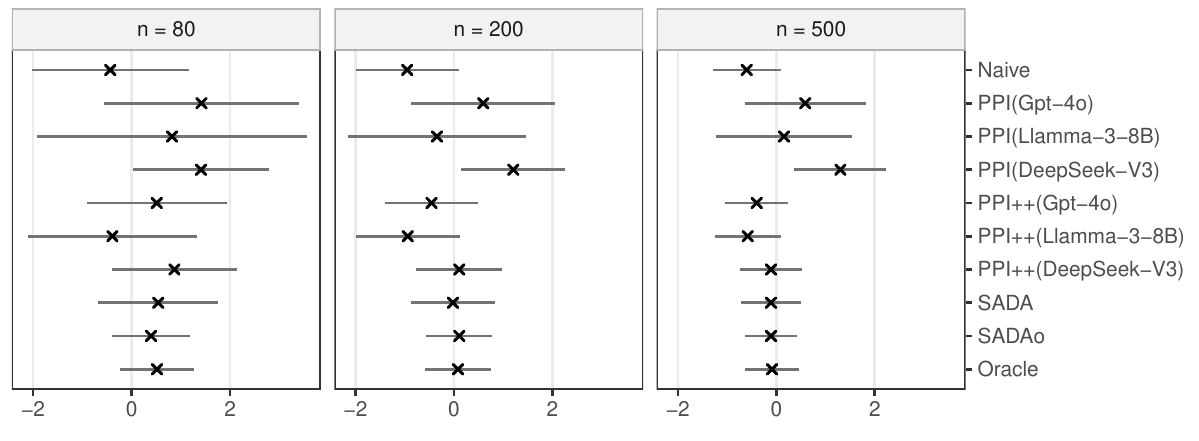}
    \caption{Point estimates and 95\% confidence intervals of different methods leveraging various prediction strategies. The estimand is the regression coefficient of politeness score on indicative modal features.}
    \label{fig:politeness_ci}
\end{figure}

\Cref{fig:politeness} reports the variation of standard deviations of different methods as the labeled data size increases, while \Cref{fig:politeness_ci} presents the corresponding point estimates and 95\% confidence intervals for $n\in\{80,200,500\}$. The three panels in \Cref{fig:politeness} differ only in the implementation of the PPI and PPI++ methods, with each panel using predictions from a different single LLM. 
While the standard deviations of all methods generally decrease with more labeled data, the PPI and PPI++ estimators do not consistently improve upon the naive approach in terms of efficiency.
In contrast, the SADA estimator, which integrates all predictions, consistently outperforms all other methods. Its performance closely tracks that of the best-performing PPI++ estimator based on GPT-4o, highlighting its adaptive behavior.
We also plot the oracle SADA estimator, SADAo, which includes the ground truth as one of its predictions. As shown, SADAo performs almost identically to the oracle benchmark obtained by regressing the ground-truth outcome on the covariates using the full data, further confirming the desired adaptivity of the proposed method.
\Cref{fig:politeness_ci} likewise shows that the SADA estimator is close to the oracle estimator and appears more stable and efficient than PPI and PPI++.

\subsection{Image Classification}\label{ssec:exp_image}
In this section, for prediction purposes, we apply SADA to the ImageNet-5 data set, which is constructed from a random subset of five classes from ImageNet-1k \citep{russakovsky2015imagenet}. The data consist of five visually diverse object categories, including animals, artifacts, and natural scenes, each paired with a ground truth class label from the human annotated ImageNet hierarchy. The data set contains 1300 images per class, for a total of 6500 images across five classes, and is split into 5000 training images, 750 validation images, and 750 test images. Among the training images, we randomly select 2500 as unlabeled data and vary the labeled-unlabeled ratio from 0.2 to 1.

We generate two pseudo labels $\wh Y_1$ and $\wh Y_2$ by training two teacher models, EfficientNet-B0 \citep{TanLe2020} and RegNetY-008 \citep{RadosavovicKosarajuGirshickHeEtAl2020}, using the following hyperparameters: 10 training epochs, a learning rate of 1e-3, a batch size of 256, and the AdamW optimizer. These trained models are then used to generate pseudo labels for all 5000 training images.

\begin{figure}
    \centering
    \includegraphics[width=0.6\textwidth]{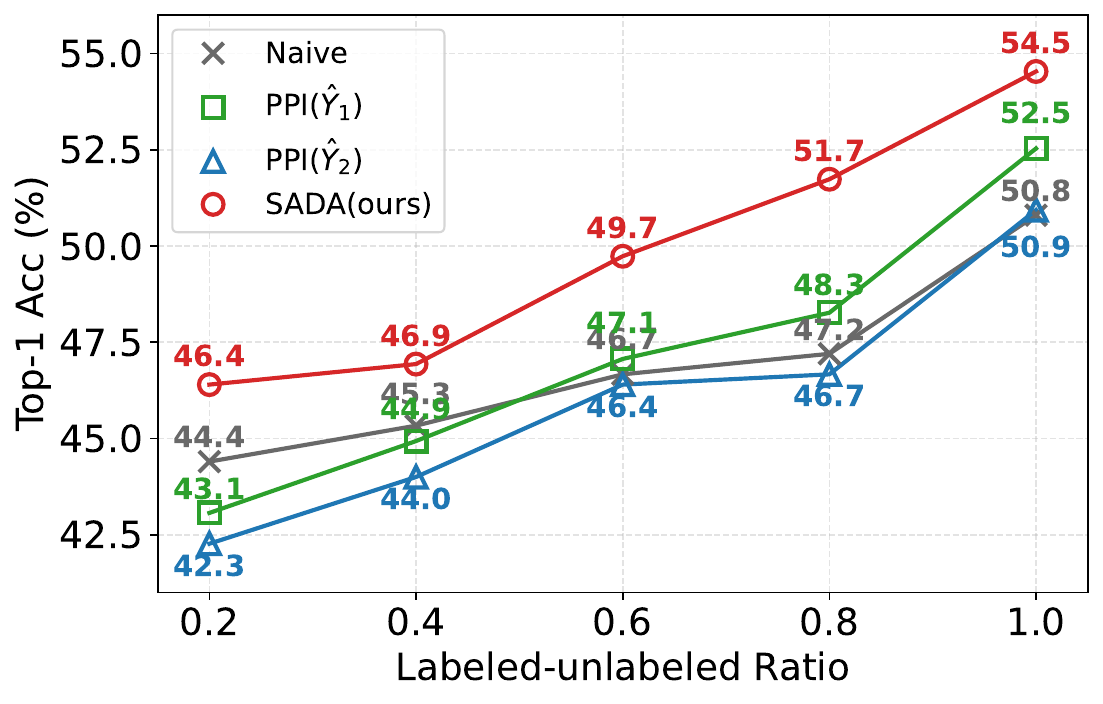}
    \caption{Top-1 accuracy of various methods on the test images across increasing numbers of labeled data.}
    \label{fig:image}
\end{figure}

We evaluate the different methods by training DaViT-T \citep{ding2022davit}, a popular vision transformer architecture that achieves state-of-the-art performance on ImageNet. The model is trained for 15 epochs with a learning rate of 1e-3, a batch size of 256, and the default AdamW optimizer. \Cref{fig:image} reports the Top-1 accuracy on the test set for the naive baseline, PPI using $\wh Y_{1}$ and $\wh Y_{2}$ separately, and SADA across labeled-unlabeled ratios ranging from 0.2 to 1. It is shown that SADA consistently achieves the best performance in terms of prediction accuracy across all settings.

\section{Discussion}\label{sec:discussion}
In this paper, we propose a novel procedure, SADA, that safely and adaptively aggregates multiple predicted labels originating from diverse state-of-the-art machine learning models for both inference and prediction purposes. An attractive feature of SADA is that it does not require any assumptions on the form, quality, or source of the predictions, which may range from highly accurate and close to the ground truth to extremely poor and essentially random noise. The algorithm automatically aggregates these predictions by assigning them appropriate weights, leading to more efficient inference or improved predictive accuracy. In addition, SADA is user-friendly, as it does not require tuning of hyperparameters or any preliminary assessment of the quality of the available predictions.

There are several directions for future exploration. An essential assumption in this paper is that the labeled and unlabeled data are drawn from the same distribution. In settings with distribution shift, the current method should be adapted accordingly, because the augmentation terms in \eqref{eq:generalEE} will not converge to zero. In such cases, distributional adjustment based on the type of shift, such as covariate shift \citep{ying2025towards} or label shift \citep{lee2025doubly,lee2025efficient} between the labeled and unlabeled data, is necessary to restore the effectiveness of the proposed methods. We are going to investigate these extensions in future work.


\section*{Supplementary material}
All technical proofs and additional experimental details are provided in the Supplement that follows. A user-friendly R package, \texttt{sada}, is available at \url{https://github.com/jw-shan/sada}.

\bibliographystyle{apalike} 
\bibliography{reference2}

\clearpage
\phantomsection
\hypertarget{supp.top}{}
\bookmark[level=0, dest=supp.top, open=false]{Supplement}
\section*{SUPPLEMENT}
\renewcommand{\thetheorem}{S\arabic{theorem}}
\renewcommand{\thelemma}{S\arabic{lemma}}
\renewcommand{\theremark}{S\arabic{remark}}
\renewcommand{\theequation}{S\arabic{equation}}
\renewcommand{\thesection}{S\arabic{section}}
\renewcommand{\thetable}{S\arabic{table}}
\def\theHsection{S\arabic{section}}
\def\theHtheorem{ST\arabic{section}.\arabic{theorem}}
\def\theHlemma{ST\arabic{section}.\arabic{theorem}}
\def\theHremark{SR\arabic{section}.\arabic{remark}}
\def\theHequation{SE\arabic{equation}}
\def\theHtable{STB\arabic{table}}
\setcounter{theorem}{0}
\setcounter{remark}{0}
\setcounter{equation}{0}
\setcounter{section}{0}
\setcounter{table}{0}

\section{Semiparametric efficiency bound}\label{sec:eif}
Here we briefly introduce the regular and asymptotically linear (RAL) estimator and influence function. In general, given i.i.d. copies of the random sample $\left\{\mathbf{z}_1, \ldots, \mathbf{z}_n\right\}$ with sample size $n$, an estimator for the parameter of interest $\bm\beta$, $\wh{\bm\beta}$, is a RAL estimator if

\begin{align*}
\sqrt{n}(\wh{\bm\beta}-\bm\beta)=n^{-1/2} \sum_{i=1}^n \bm\phi\left(\mathbf{z}_i\right)+o_p(1),
\end{align*}
where the zero-mean function $\bm\phi(\cdot)$ is called the influence function of $\wh{\bm\beta}$. Then, the central limit theorem implies that $\sqrt{n}(\wh{\bm\beta}-\bm\beta)\convd \calN(\0,\mE(\bm\phi \bm\phi\trans))$, provided that $\mE(\bm\phi \bm\phi\trans)$ is finite and nonsingular. 
Among all RAL estimators for $\bm\beta$, the influence function of the one with the smallest asymptotic variance is called the efficient influence function (EIF), $\bm\phi_\eif$, and the semiparametric efficiency bound is $\mE\left(\bm\phi_\eif \bm\phi_\eif\trans\right)$.

We provide a brief, non-technical overview of the derivation of influence functions. At first, the likelihood can be decomposed into a parametric component and one or more nonparametric components. For each nonparametric component, we can identify a corresponding nuisance tangent space. The overall nuisance tangent space for the semiparametric model is then obtained by combining the individual nuisance tangent spaces. A valid influence function must lie in the orthogonal complement of this nuisance tangent space. Under suitable regularity conditions, an element from this perpendicular space can be chosen as the influence function.

We refer readers to \citet{BickelKlaassenRitovWellner1993} and \citet{Tsiatis2006} for further interpretations. In addition, \citet{Kennedy2023} provided several strategies for deriving the efficient influence function together with many examples.

As an illustration, consider the outcome mean $\theta^*=\mE(Y)$ discussed in \Cref{ssec:mean_estimation}. Denote $r_i=1$ for labeled units $i=1,\dots,n$ and 0 for unlabeled units, and suppose $n/N\to \pi\in(0,1)$ as $n,N\to\infty$ and $\wh Y_k=\wh f_k(\X)$ for $k=1,\dots,K$. Then, based on labeled data $\calL=\{(\x_i,y_i),i=1,\dots,n\}$ and unlabeled data $\calU=\{\x_i,i=n+1,\dots,N\}$, the EIF for $\theta^*$ takes the form
\begin{align*}
  \phi_\eif(r,\x,y) = r \pi^{-1} \{y-\mE(Y\mid\x)\} + \mE(Y\mid\x) -\theta^*,
\end{align*}
which is a special case of \Cref{thm:adaptivity}(ii) applied to the score $\s(\x,y;\theta)=y-\theta$.

\section{Proof of Proposition~\ref{prop:opt_weight}}\label{sec:app_pf_prop_opt_weight}
\begin{proof}
  Recall that  $\wh\btt(\calW)$ solves 
  \begin{align*}
    \frac{1}{n}\sum_{i=1}^{n} \s(\x_i, y_i;\btt)+  \calW\trans \left\{\frac{1}{N-n}  \sum_{i=n+1}^{N} \calS(\x_i,\wh\y_{i};\btt) - \frac{1}{n} \sum_{i=1}^{n} \calS(\x_i,\wh\y_{i};\btt)\right\} =\0,
  \end{align*}
  where  $\calW=(\calW_1\trans,\calW_2\trans,\dots,\calW_K\trans)\trans\in\mR^{(Kp)\times p}$, and $\calS(\x,\wh\y;\btt)= (\s(\x,\wh y_1;\btt)\trans,\cdots,\s(\x,\wh y_K;\btt)\trans)\trans$.
  Then the standard Taylor expansion yields that
  \begin{align*}
    \0=&~  \frac{1}{n}\sum_{i=1}^{n} \s(\x_i, y_i;\btt^*)+  \calW\trans \left\{\frac{1}{N-n}  \sum_{i=n+1}^{N} \calS(\x_i,\wh\y_{i};\btt^*) - \frac{1}{n} \sum_{i=1}^{n} \calS(\x_i,\wh\y_{i};\btt^*)\right\} \\ 
    &~ + \left[\frac{1}{n}\sum_{i=1}^{n} \frac{\partial}{\partial\btt\trans} \s(\X, Y;\overline\btt)+  \calW\trans \left\{\frac{1}{N-n}  \sum_{i=n+1}^{N} \frac{\partial}{\partial\btt\trans}\calS(\x_i,\wh\y_{i};\overline\btt) - \frac{1}{n} \sum_{i=1}^{n} \frac{\partial}{\partial\btt\trans}\calS(\x_i,\wh\y_{i};\overline\btt)\right\}\right] \\ 
    & \hspace{11.5cm}\times\{\wh\btt(\calW)-\btt^*\},
  \end{align*}
  where $\overline\btt$ lies between $\wh\btt(\calW)$ and $\btt^*$. Then following uniform weak law of large number \citep{NeweyMcFadden1994}, under the regularity conditions, we have 
  \begin{align*}
    \wh\btt(\calW)-\btt^* \doteq -\H^{-1} \left[ \frac{1}{n}\sum_{i=1}^{n} \s(\x_i, y_i;\btt^*)+  \calW\trans \left\{\frac{1}{N-n}  \sum_{i=n+1}^{N} \calS(\x_i,\wh\y_{i};\btt^*) - \frac{1}{n} \sum_{i=1}^{n} \calS(\x_i,\wh\y_{i};\btt^*)\right\}\right],
  \end{align*}
  where $\H=\mE\{\partial \s(\X, Y;\btt^*)\trans/\partial\btt\}$.
  Then 
  \begin{align*}
  \mE[\{\wh\btt(\calW) - &\btt^*\}^{\otimes 2}] 
  \doteq  \H^{-1}\Bigg[ \frac{1}{n}\var\{\s(\X,Y;\btt^*)\} + \frac{N}{n(N-n)}\calW\trans \var\{\calS(\X,\wh\Y;\btt^*)\}\calW   \\ 
  & -  \frac{1}{n}  \mE\{\s(\X,Y;\btt^*)\calS(\X,\wh\Y;\btt^*)\trans\} \calW
  - \frac{1}{n}  \calW\trans \mE\{\calS(\X,\wh\Y;\btt^*)\s(\X,Y;\btt^*)\trans\}\Bigg]\H^{-\T}.
  \end{align*}
  Let 
  \begin{align*}
  \calW^\opt = \frac{N-n}{N} \var\{\calS(\X,\wh\Y;\btt^*)\}^{-1} \mE\{\calS(\X,\wh\Y;\btt^*)\s(\X,Y;\btt^*)\trans\}.
  \end{align*}
  We next show that 
  \begin{align}\label{eq:app_pf_EE_T1}
    \mE[\{\wh\btt(\calW^\opt) - \btt^*\}^{\otimes 2}] \preceq \mE[\{\wh\btt(\calW) - \btt^*\}^{\otimes 2}],~~\forall\calW\in\mR^{(Kp)\times p},
  \end{align}
  i.e., $\mE[\{\wh\btt(\calW) - \btt^*\}^{\otimes 2}]-\mE[\{\wh\btt(\calW^\opt) - \btt^*\}^{\otimes 2}]$ is a positive semi-definite matrix for any $\calW\in\mR^{(Kp)\times p}$.
  Note that 
  \begin{align*}
    \mE[\{\wh\btt(\calW^\opt) - \btt^*\}^{\otimes 2}] 
  =&~  \H^{-1}\Bigg[ \frac{1}{n}\var\{\s(\X,Y;\btt^*)\}  - \left(\frac{1}{n}-\frac{1}{N}\right) \mE\{\s(\X,Y;\btt^*)\calS(\X,\wh\Y;\btt^*)\trans\}\\
   &\qquad\qquad \times \var\{\calS(\X,\wh\Y;\btt^*)\}^{-1}
   \mE\{\calS(\X,\wh\Y;\btt^*)\s(\X,Y;\btt^*)\trans\}\Bigg]\H^{-\T}.
  \end{align*}
  For any non-zero vector $\a\in\mR^p$, let $\wt\a=\H^{-\T}\a$, then we have 
  \begin{align*}
  & \a\trans \left(\mE[\{\wh\btt(\calW) - \btt^*\}^{\otimes 2}]-\mE[\{\wh\btt(\calW^\opt) - \btt^*\}^{\otimes 2}]\right)\a \\ 
  =&~ \frac{N}{n(N-n)} \left( \calW\wt\a - \frac{N-n}{N}\var\{\calS(\X,\wh\Y;\btt^*)\}^{-1} \mE\{\calS(\X,\wh\Y;\btt^*)\s(\X,Y;\btt^*)\trans\}\right)\trans \\ 
  &\qquad\qquad \times \var\{\calS(\X,\wh\Y;\btt^*)\} \\ 
  &\qquad\qquad \times \left( \calW\wt\a - \frac{N-n}{N}\var\{\calS(\X,\wh\Y;\btt^*)\}^{-1} \mE\{\calS(\X,\wh\Y;\btt^*)\s(\X,Y;\btt^*)\trans\}\right) \\ 
  \ge&~ 0.
  \end{align*}
  Therefore, \eqref{eq:app_pf_EE_T1} holds by definition.
\end{proof}

\section{Adaptivity in the mean estimation example}\label{sec:app_supp_mean}
Following \Cref{ssec:mean_estimation}, we derive the optimal weight $\bm\omega^\opt$ in \eqref{eq:w_opt_mean} and the resulting form of $\wh\theta^\sada$ under two scenarios: (i) $\wh Y_1\equiv Y$, and (ii) $\wh Y_k=\wh f_k(\X)$  for $k=1,\dots,K$, with $\wh Y_1\equiv\mE(Y\mid\X)$.

\paragraph*{Case 1: $\wh Y_1\equiv Y$}
   Denote $\wh\Y_{-1}=(\wh\Y_2,\dots,\wh\Y_K)\trans$, and partition
  \begin{align*}
    \var(\wh\Y) =
    \left(\begin{array}{cc}
      \var(Y) & \cov(\wh\Y_{-1},Y)\trans  \\
      \cov(\wh\Y_{-1},Y)  & \var(\wh\Y_{-1})
      \end{array}\right)
      =:
    \left(\begin{array}{cc}
      C_{11} & \bm C_{21}\trans  \\
      \bm C_{21}  & \bm C_{22}
      \end{array}\right).
  \end{align*}
  By the inversion formula of $2\times 2$ block matrix \citep[Theorem 2.1]{LuShiou2002},
      \begin{align}\label{eq:app_inverse}
    \var(\wh\Y)^{-1}=
    \left(\begin{matrix}
    C_{11}^{-1}+ C_{11}^{-1} \bm C_{21}\bm M^{-1} \bm C_{21}\trans C_{11}^{-1} & - C_{11}^{-1} \bm C_{21}\bm M^{-1} \\
    -\bm M^{-1} \bm C_{21}\trans C_{11}^{-1} & \bm M^{-1}
    \end{matrix}\right),
  \end{align}
  where $\bm M=\bm C_{22}-\bm C_{21}\trans C_{11}^{-1}\bm C_{21}$, and $\cov(\wh\Y,Y)=(C_{11},\bm C_{21}\trans)\trans$.
  Substituting these expressions into \eqref{eq:w_opt_mean} yields
  \begin{align*}
    \bm\omega^\opt
    = \frac{N-n}{N} \var(\wh\Y)^{-1}\cov(\wh\Y,Y)
    = \frac{N-n}{N} (1,0,\dots,0)\trans.
  \end{align*}
  Consequently, the SADA estimator reduces to $\wh\theta^\sada=N^{-1}\sum_{i=1}^N \wh y_{1,i}$, with mean squared error
  \begin{align*}
  \mE \{(\wh\theta^\sada-\theta^*)^2\} = \frac{1}{N} \var(Y),
  \end{align*}
  which matches the variance of the oracle estimator that has access to the true outcomes on the unlabeled units.

\paragraph*{Case 2: $\wh Y_k = \wh f_k(\X)$ for $k=1,\dots,K$, and $\wh Y_1\equiv \mE(Y\mid\X)$}
  Let $\wh\f_{-1}(\x)=(\wh f_2(\x),\dots,\wh f_K(\x))\trans$. Then,
  \begin{align*}
    \var(\wh\Y) =
    \left(\begin{array}{cc}
      \var\{\mE(Y\mid\X)\} & \cov\{\wh\f_{-1}(\X),\mE(Y\mid\X)\}\trans  \\
      \cov\{\wh\f_{-1}(\X),\mE(Y\mid\X)\}  & \var\{\wh\f_{-1}(\X)\}
      \end{array}\right)
      =:
    \left(\begin{array}{cc}
      D_{11} & \bm D_{21}\trans  \\
      \bm D_{21}  & \bm D_{22}
      \end{array}\right).
  \end{align*}
  Moreover, because $\wh\f_{-1}(\X)$ is a measurable function of $\X$, the tower property of conditional expectation gives
  \begin{align*}
  \cov(\wh\Y,Y)
  =\cov\left(
  \left(\begin{aligned}
    \mE(Y\mid\X)\\
    \wh\f_{-1}(\X)
  \end{aligned} \right)
  ,Y\right)
  =\cov\left(
  \left(\begin{aligned}
    \mE(Y\mid\X)\\
    \wh\f_{-1}(\X)
  \end{aligned} \right)
  ,\mE(Y\mid\X)\right)
  =:\left(\begin{aligned}
    D_{11}\\
    \bm D_{21}
  \end{aligned} \right).
  \end{align*}
  Applying the inversion formula of $2\times 2$ block matrix as in \eqref{eq:app_inverse}, we obtain
  \begin{align*}
    \bm\omega^\opt
    = \frac{N-n}{N} \var(\wh\Y)^{-1}\cov(\wh\Y,Y)
    = \frac{N-n}{N} (1,0,\dots,0)\trans.
  \end{align*}
  Therefore, the SADA estimator simplifies to
  \begin{align*}
  \wh\theta^\sada = \frac{1}{n}\sum_{i=1}^{n}\{y_i - \mE(Y\mid\x_i)\} + \frac{1}{N}\sum_{i=1}^{N}\mE(Y\mid\x_i).
  \end{align*}
  Let $r_i=1$ for labeled units and $r_i=0$ otherwise, and assume $n/N\to\pi\in(0,1)$ as $n,N\to\infty$. Then,
  \begin{align*}
    \sqrt{N}(\wh\theta^\sada-\theta^*)
    &= \frac{\sqrt{N}}{n}\sum_{i=1}^{n}  \{y_i - \mE(Y\mid\x_i)\} +  \frac{1}{\sqrt{N}}\sum_{i=1}^{N} \{\mE(Y\mid\x_i) - \theta^*\}  \\
    &=\frac{1}{\sqrt{N}}\sum_{i=1}^{N} \phi_\eif(r_i,\x_i,y_i)+o_p(1),
  \end{align*}
  where
  \begin{align*}
    \phi_\eif(r,\x,y) = \frac{r}{\pi}\{y-\mE(Y\mid\x)\} + \mE(Y\mid\x) -\theta^*
  \end{align*}
  is the efficient influence function (EIF) for estimating $\theta^*$ based on the labeled data $\calL$ and unlabeled data $\calU$; this expression coincides with the general form derived in \Cref{thm:adaptivity}(ii) specialized to the score $\s(\x,y;\theta)=y-\theta$. By the central limit theorem, $\sqrt{N}(\wh\theta^\sada-\theta^*)\convd\calN\{0,\mE(\phi_\eif^2)\}$, so $\wh\theta^\sada$ attains the semiparametric efficiency bound.

\section{Proof of Theorem~\ref{thm:general_eff_gain}}\label{sec:app_pf_thm_eff_gain}

We first present the following lemma, which establishes the consistency of $\wh\calW^\opt$ defined in Section \ref{sec:sada_inference} for $\calW^\opt$.

\begin{lemma}\label{lem:Wopt-consistency}
Suppose $\wh\btt\convp\btt^*$ and Assumption \ref{ass:regularity} holds, then $\wh\calW^\opt \convp \calW^\opt$.
\end{lemma}

\begin{proof}
Denote $\calS_i(\btt)=\calS(\X_i,\wh\Y_i;\btt)$ and $\s_i(\btt)=\s(\X_i,Y_i;\btt)$.
Since $K$ is fixed, Assumption~\ref{ass:regularity}(iii)--(iv) implies that,
there exists $\Theta_0=\{\btt:\|\btt-\btt^*\|<\delta\}\subset \Theta$ such that
\begin{align*}
\mE\left\{\sup_{\btt\in\Theta_0}\|\calS(\X,\wh\Y;\btt)\|^2\right\}<\infty,
~~
\mE\left\{\sup_{\btt\in\Theta_0}\left\|
\frac{\partial \calS(\X,\wh\Y;\btt)\trans}{\partial\btt}\right\|^2\right\}<\infty,
\end{align*}
and
\begin{align*}
\mE\left\{\sup_{\btt\in\Theta_0}\|\s(\X,Y;\btt)\|^2\right\}<\infty,
~~
\mE\left\{\sup_{\btt\in\Theta_0}\left\|
\frac{\partial \s(\X,Y;\btt)\trans}{\partial\btt}\right\|^2\right\}<\infty.
\end{align*}
Then, the mean value theorem yields that for $\btt\in\Theta_0$,
\begin{align*}
\|\calS_i(\btt)-\calS_i(\btt^*)\|
&\le
\sup_{\btt\in\Theta_0}\left\|
\frac{\partial \calS_i(\btt)\trans}{\partial\btt}\right\|
\,\|\btt-\btt^*\|,
\\
\|\s_i(\btt)-\s_i(\btt^*)\|
&\le
\sup_{\btt\in\Theta_0}\left\|
\frac{\partial \s_i(\btt)\trans}{\partial\btt}\right\|
\,\|\btt-\btt^*\|.
\end{align*}

We first show that
\begin{align*}
\bar\calS_N(\wh\btt):=\frac1N\sum_{i=1}^N \calS_i(\wh\btt)
\convp
\mE\{\calS(\X,\wh\Y;\btt^*)\}.
\end{align*}
Note that
\begin{align*}
\left\|
\frac1N\sum_{i=1}^N\calS_i(\wh\btt)-\frac1N\sum_{i=1}^N\calS_i(\btt^*)
\right\|
&\le
\|\wh\btt-\btt^*\|
\cdot
\frac1N\sum_{i=1}^N
\sup_{\btt\in\Theta_0}\left\|
\frac{\partial \calS_i(\btt)\trans}{\partial\btt}\right\|.
\end{align*}
By the weak law of large numbers, under Assumption~\ref{ass:regularity}, the sample average on the right converges in
probability to a finite constant, while $\wh\btt-\btt^*=o_p(1)$. Hence
\begin{align*}
\frac1N\sum_{i=1}^N\calS_i(\wh\btt)
-
\frac1N\sum_{i=1}^N\calS_i(\btt^*)
=o_p(1).
\end{align*}
Since $\mE\|\calS(\X,\wh\Y;\btt^*)\|<\infty$, the weak law of large numbers also
gives
\begin{align*}
\frac1N\sum_{i=1}^N\calS_i(\btt^*)
\convp
\mE\{\calS(\X,\wh\Y;\btt^*)\},
\end{align*}
and therefore
\begin{align*}
\bar\calS_N(\wh\btt)\convp\mE\{\calS(\X,\wh\Y;\btt^*)\}.
\end{align*}

Next consider
\begin{align*}
\frac1N\sum_{i=1}^N \calS_i(\wh\btt)\calS_i(\wh\btt)\trans.
\end{align*}
Using the fact $aa\trans-bb\trans=(a-b)a\trans+b(a-b)\trans$, we obtain
\begin{align*}
&\left\|
\frac1N\sum_{i=1}^N\calS_i(\wh\btt)\calS_i(\wh\btt)\trans
-
\frac1N\sum_{i=1}^N\calS_i(\btt^*)\calS_i(\btt^*)\trans
\right\| \\
&\le
\frac1N\sum_{i=1}^N
\|\calS_i(\wh\btt)-\calS_i(\btt^*)\|\,\|\calS_i(\wh\btt)\|
+
\frac1N\sum_{i=1}^N
\|\calS_i(\btt^*)\|\,\|\calS_i(\wh\btt)-\calS_i(\btt^*)\| \\
&\le
\|\wh\btt-\btt^*\|
\left[
\frac1N\sum_{i=1}^N
\sup_{\btt\in\Theta_0}\left\|
\frac{\partial \calS_i(\btt)\trans}{\partial\btt}\right\|
\sup_{\btt\in\Theta_0}\|\calS_i(\btt)\|
+
\frac1N\sum_{i=1}^N
\sup_{\btt\in\Theta_0}\left\|
\frac{\partial \calS_i(\btt)\trans}{\partial\btt}\right\|
\|\calS_i(\btt^*)\|
\right].
\end{align*}
By the Cauchy--Schwarz inequality, under Assumption~\ref{ass:regularity},
\begin{align*}
\frac1N\sum_{i=1}^N
\sup_{\btt\in\Theta_0}\left\|
\frac{\partial \calS_i(\btt)\trans}{\partial\btt}\right\|
\sup_{\btt\in\Theta_0}\|\calS_i(\btt)\|
&\le
\left\{
\frac1N\sum_{i=1}^N
\sup_{\btt\in\Theta_0}\left\|
\frac{\partial \calS_i(\btt)\trans}{\partial\btt}\right\|^2
\right\}^{1/2}
\left\{
\frac1N\sum_{i=1}^N
\sup_{\btt\in\Theta_0}\|\calS_i(\btt)\|^2
\right\}^{1/2}
\\
&=O_p(1),
\end{align*}
and similarly,
\begin{align*}
\frac1N\sum_{i=1}^N
\sup_{\btt\in\Theta_0}\left\|
\frac{\partial \calS_i(\btt)\trans}{\partial\btt}\right\|
\|\calS_i(\btt^*)\|
=O_p(1).
\end{align*}
Therefore,
\begin{align*}
\frac1N\sum_{i=1}^N\calS_i(\wh\btt)\calS_i(\wh\btt)\trans
-
\frac1N\sum_{i=1}^N\calS_i(\btt^*)\calS_i(\btt^*)\trans
=o_p(1).
\end{align*}
Since $\mE\|\calS(\X,\wh\Y;\btt^*)\|^2<\infty$, the weak law of large numbers
implies
\begin{align*}
\frac1N\sum_{i=1}^N\calS_i(\btt^*)\calS_i(\btt^*)\trans
\convp
\mE\{\calS(\X,\wh\Y;\btt^*)\calS(\X,\wh\Y;\btt^*)\trans\}.
\end{align*}
Hence
\begin{align*}
\frac1N\sum_{i=1}^N\calS_i(\wh\btt)\calS_i(\wh\btt)\trans
\convp
\mE\{\calS(\X,\wh\Y;\btt^*)\calS(\X,\wh\Y;\btt^*)\trans\}.
\end{align*}
Similarly, we can show that
\begin{align}\label{eq:app_pf_consistency1}
\frac1n\sum_{i=1}^n \calS_i(\wh\btt)\s_i(\wh\btt)\trans
\convp
\mE\{\calS(\X,\wh\Y;\btt^*)\s(\X,Y;\btt^*)\trans\}.
\end{align}

Combining the above results yields that
\begin{align*}
\frac1N\sum_{i=1}^N
\left\{\calS_i(\wh\btt)-\bar\calS_N(\wh\btt)\right\}^{\otimes2}
=
\frac1N\sum_{i=1}^N \calS_i(\wh\btt)\calS_i(\wh\btt)\trans
-\bar\calS_N(\wh\btt)\bar\calS_N(\wh\btt)\trans 
\convp
\var\{\calS(\X,\wh\Y;\btt^*)\}.
\end{align*}
Since $\var\{\calS(\X,\wh\Y;\btt^*)\}$ is nonsingular, the continuous mapping
theorem implies that
\begin{align}\label{eq:app_pf_consistency2}
\left[
\frac1N\sum_{i=1}^N
\left\{\calS_i(\wh\btt)-\bar\calS_N(\wh\btt)\right\}^{\otimes2}
\right]^{-1}
\convp
\var\{\calS(\X,\wh\Y;\btt^*)\}^{-1}.
\end{align}
Combining \eqref{eq:app_pf_consistency1} and \eqref{eq:app_pf_consistency2}, the Slutsky's theorem implies that
\begin{align*}
\wh\calW^\opt
&=
\frac{N-n}{N}
\left[
\frac1N\sum_{i=1}^N
\left\{\calS_i(\wh\btt)-\bar\calS_N(\wh\btt)\right\}^{\otimes2}
\right]^{-1}
\left\{
\frac1n\sum_{i=1}^n \calS_i(\wh\btt)\s_i(\wh\btt)\trans
\right\} \convp
\calW^\opt.
\end{align*}
This completes the proof.
\end{proof}

\begin{proof}[Proof of Theorem~\ref{thm:general_eff_gain}]
  Following arguments similar to those in the proof of \Cref{prop:opt_weight} in \Cref{sec:app_pf_prop_opt_weight}, under \cref{ass:regularity}, we have 
  \begin{small}
    \begin{equation}\label{eq:app_pf_thm_gain_1}
        \begin{aligned}
          &\wh\btt^\sada-\btt^* \\ 
          &~\doteq -\H^{-1} \left[ \frac{1}{n}\sum_{i=1}^{n} \s(\x_i, y_i;\btt^*)+  (\wh\calW^\opt)\trans \left\{\frac{1}{N-n}  \sum_{i=n+1}^{N} \calS(\x_i,\wh\y_{i};\btt^*) - \frac{1}{n} \sum_{i=1}^{n} \calS(\x_i,\wh\y_{i};\btt^*)\right\}\right] \\ 
          &~= -\H^{-1} \left[ \frac{1}{n}\sum_{i=1}^{n} \s(\x_i, y_i;\btt^*)+  (\calW^\opt)\trans \left\{\frac{1}{N-n}  \sum_{i=n+1}^{N} \calS(\x_i,\wh\y_{i};\btt^*) - \frac{1}{n} \sum_{i=1}^{n} \calS(\x_i,\wh\y_{i};\btt^*)\right\}\right] \\ 
          &\quad - \H^{-1}  (\wh\calW^\opt-\calW^\opt)\trans \left\{\frac{1}{N-n}  \sum_{i=n+1}^{N} \calS(\x_i,\wh\y_{i};\btt^*) - \frac{1}{n} \sum_{i=1}^{n} \calS(\x_i,\wh\y_{i};\btt^*)\right\}\\
          &~ =: T_1+T_2,
        \end{aligned}
    \end{equation}
  \end{small}
  where $\H=\mE\{\partial \s(\X, Y;\btt^*)\trans/\partial\btt\}$. 
  We next show that $T_2=o_p(n^{-1/2})$. Recall that 
  \begin{align*}
    \calW^\opt = \frac{N-n}{N} \var\{\calS(\X,\wh\Y;\btt^*)\}^{-1} \mE\{\calS(\X,\wh\Y;\btt^*)\s(\X,Y;\btt^*)\trans\}=:\frac{N-n}{N}\overline\calW.
  \end{align*}
  Then,
  \begin{align*}
  \|T_2\| \le &~  \left\|\H^{-1} \right\| \times \frac{N-n}{N} \left\|\wh{\overline\calW}-\overline\calW\right\| \times \left\|\frac{1}{N-n}  \sum_{i=n+1}^{N} \calS(\x_i,\wh\y_{i};\btt^*) - \frac{1}{n} \sum_{i=1}^{n} \calS(\x_i,\wh\y_{i};\btt^*)\right\| \\ 
  =&~ O(1)\times o_p\left(\frac{N-n}{N}\right) \times O_p \left(\sqrt{\frac{1}{N-n}+\frac{1}{n}}\right) = o_p(n^{-1/2}),
  \end{align*}
  where the first equality is by $\wh\calW^\opt\convp\calW^\opt$, Chebyshev's inequality and
  \begin{align*}
    \mE\left\|\frac{1}{N-n}  \sum_{i=n+1}^{N} \calS(\x_i,\wh\y_{i};\btt^*) - \frac{1}{n} \sum_{i=1}^{n} \calS(\x_i,\wh\y_{i};\btt^*)\right\|^2 = \left(\frac{1}{N-n}+\frac{1}{n}\right)\left\|\var\{\calS(\X,\wh\Y;\btt^*)\}\right\|.
  \end{align*}
  Moreover,
  \begin{align*}
  \var(T_1)=&~ \H^{-1} \Bigg[
  \frac{1}{n} \var\{\s(\x_i, y_i;\btt^*)\} +  \left(\frac{1}{N-n}+\frac{1}{n}\right)(\calW^\opt)\trans\var\{\calS(\X,\wh\Y;\btt^*)\}\calW^\opt \\ 
  &\qquad\qquad - \frac{2}{n} \cov\{\s(\x_i, y_i;\btt^*),\calS(\X,\wh\Y;\btt^*)\}\calW^\opt
  \Bigg]\H^{-\T}\\ 
  =&~  \H^{-1}\left\{\frac{1}{n}\bSig_\nv-\frac{N-n}{Nn}\bSig_g\right\}\H^{-\T},
  \end{align*}
  where $\bSig_\nv=\var\{\s(\x_i, y_i;\btt^*)\}$ and
  \begin{align*}
      \bSig_g =  \mE\{\s(\X,Y;\btt^*)\calS(\X,\wh\Y;\btt^*)\trans\} \var\{\calS(\X,\wh\Y;\btt^*)\}^{-1}
  \mE\{\calS(\X,\wh\Y;\btt^*)\s(\X,Y;\btt^*)\trans\}.
  \end{align*}
  By the central limit theorem and the Slutsky's theorem, we have
  \begin{align*}
    \sqrt{n}(\wh\btt^\sada - \btt^*)\convd \calN\Big(\0,\H^{-1}\big\{\bSig_\nv-\frac{N-n}{N}\bSig_g\big\}\H^{-\T}\Big).
  \end{align*}

\end{proof}

\section{Proof of Theorem~\ref{thm:adaptivity}}\label{sec:app_pf_thm_adap}
\begin{proof}
  \noindent\emph{Part (i).}
  Without loss of generality, we assume $\wh Y_1\equiv Y$. We denote $\s\equiv \s_1(\X,\wh Y_1;\btt^*)\equiv \s(\X,Y;\btt^*)$ and $\s_{-1}\equiv (\s(\X,\wh Y_2;\btt^*)\trans,\cdots,\s(\X,\wh Y_K;\btt^*)\trans)\trans$. Then 
  \begin{align*}
    \var\{\calS(\X,\wh\Y;\btt^*)\} =  
    \left(\begin{array}{cc}
      \var(\s_1) & \cov(\s_{-1},\s_1)\trans  \\
      \cov(\s_{-1},\s_1)  & \var(\s_{-1})
      \end{array}\right),
  \end{align*}
  and $\mE\{\calS(\X,\wh\Y;\btt^*)\s(\X,Y;\btt^*)\trans\}=(\var(\s_1),\cov(\s_{-1},\s_1)\trans)\trans$. Similar to the proof in \Cref{sec:app_supp_mean}, by the inversion formula of block matrix, we have $\calW^\opt=(\I,\0,\dots,\0)\trans\cdot (N-n)/N$, and
    \begin{align*}
        \bSig_g =&~   \mE\{\s(\X,Y;\btt^*)\calS(\X,\wh\Y;\btt^*)\trans\} \var\{\calS(\X,\wh\Y;\btt^*)\}^{-1}
   \mE\{\calS(\X,\wh\Y;\btt^*)\s(\X,Y;\btt^*)\trans\} \\ 
    =&~  \var(\s_1) = \bSig_\nv.
    \end{align*}
  Then by  \Cref{thm:general_eff_gain}, we have 
  \begin{align*}
    \wh\btt^\sada - \btt^*\convd \calN\left(\0,\frac1N\H^{-1}\bSig_\nv\H^{-\T}\right).
  \end{align*}
  It is asymptotically equivalent to the oracle estimator who knows the ground truth of the unlabeled data and solves
  \begin{align*}
  \frac{1}{N}\sum_{i=1}^{N} \s(\x_i,y_i;\btt)=0.
  \end{align*}

  \noindent\emph{Part (ii).}
  We first derive the EIF for estimating $\btt^*$, based on labeled data $\calL=\{(\x_i,y_i),i=1,\dots,n\}$ and unlabeled data $\calU=\{\x_i,i=n+1,\dots,N\}$. The joint density from one observation is
  \begin{align*}
   \{\pi f(\x) f(y\mid\x)\}^r \{(1-\pi)f( \x)\}^{1-r}=\pi^r(1-\pi)^{1-r}f(\x)f(y\mid\x)^r.
  \end{align*} 
  It's straightforward to show the tangent space of this model is $\calT=\Lambda_1\bigoplus\Lambda_2$, where 
  \begin{align*}
    \Lambda_1 = \left\{r\b(y,\x): \mE(\b\mid \x)=0\right\}, 
    ~~\text{and}~~
    \Lambda_2 =  \left\{\a(\x): \mE(\a)=0\right\}.
  \end{align*}
  Let
  \begin{align*}
    \bm\Phi_\eif(r,\x,y) = -\H^{-1}\left[\frac{r}{\pi}\{\s(\x,y;\btt^*)-\bm\mu(\x)\} + \bm\mu(\x)\right],
  \end{align*}
  where $\bm\mu(\x)=\mE\{\s(\x,Y;\btt^*)\mid\x\}$, $\frac{r}{\pi}\{\s(\x,y;\btt^*)-\bm\mu(\x)\}\in\Lambda_1$ and $\bm\mu(\x)\in\Lambda_2$. It is easy to check $\bm\Phi_\eif$ satisfies the orthogonal conditions \citep[Theorems 4.2 and 4.3]{Tsiatis2006} and is thus the efficient influence function.

  Without loss of generality, we assume  $\s(\x,\wh y_1;\btt^*) =\s(\x,\wh f_1(\x);\btt^*)\equiv \bm\mu(\x)$. We denote $\s_{-1}(\x)=(\s(\x,\wh f_2(\x);\btt^*)\trans,\dots,\s(\x,\wh f_K(\x);\btt^*)\trans)\trans$. Then 
  \begin{align*}
    \var\{\calS(\X,\wh\Y;\btt^*)\} =  
    \left(\begin{array}{cc}
      \var\{\bm\mu(\X)\} & \cov\{\s_{-1}(\X),\bm\mu(\X)\}\trans  \\
      \cov\{\s_{-1}(\X),\bm\mu(\X)\}  & \var\{\s_{-1}(\X)\}
      \end{array}\right),
  \end{align*}
  and $\mE\{\calS(\X,\wh\Y;\btt^*)\s(\X,Y;\btt^*)\trans\}=(\var\{\bm\mu(\X)\},\cov\{\s_{-1}(\X),\bm\mu(\X)\}\trans)\trans$. Similar to the proof in \Cref{sec:app_supp_mean}, by the inversion formula of block matrix, we have $\calW^\opt=(\I,\0,\dots,\0)\trans\cdot (N-n)/N$. Then, by \eqref{eq:app_pf_thm_gain_1}, we have
  \begin{small}
    \begin{align*}
      \sqrt{N}&\{\wh\btt^\sada-\btt^*\}\\
      =& - \H^{-1} \left[ \frac{1}{n}\sum_{i=1}^{n} \s(\x_i, y_i;\btt^*)+  \frac{N-n}{N} \left\{\frac{1}{N-n}  \sum_{i=n+1}^{N} \s(\x_i,\wh y_{1,i};\btt^*) - \frac{1}{n} \sum_{i=1}^{n} \s(\x_i,\wh y_{1,i};\btt^*)\right\}\right] \\
      =&  - \H^{-1} \left[\frac{\sqrt{N}}{n}\sum_{i=1}^{n}  \{\s(\x_i,y_i;\btt^*) - \bm\mu(\x_i)\} +  \frac{1}{\sqrt{N}}\sum_{i=1}^{N} \bm\mu(\x_i)\right] +o_p(1)  \\
      =& \frac{1}{\sqrt{N}}\sum_{i=1}^{N} \bm\Phi_\eif(r_i,\x_i,y_i)+o_p(1),
    \end{align*} 
  \end{small}
  due to $n/N\to\pi$.

\end{proof}

\section{Proof of Theorem~\ref{thm:excess_risk}}\label{sec:supp_pf_thm_prediction}

\begin{proof}
  Theorem~\ref{thm:excess_risk} follows directly from theorem~\ref{thm:excess_risk_supp}, which provides a non-asymptotic bound for the expected loss.
  \end{proof}

  \begin{theorem}\label{theorem:formal}\label{thm:excess_risk_supp}
      (i) Under \Cref{ass:complexity}, we have that with probability at least $1-\delta$,
      \begin{align*}
      \calR(\wh{\btt}^\nv) -\calR(\btt^*)\leq   \frac{16\sqrt{2}C}{\sqrt{n}} \int_0^\infty \sqrt{\log N(\epsilon, \Theta, d_1)}d\epsilon+\sqrt{\frac{24B^2}{n}\log \frac{2}{\delta}}+ \frac{2B}{3n} \log \frac{2}{\delta},
  \end{align*}
  and
  \begin{align*}
       \calR(\wh \btt^\sada) - &\calR(\btt^*)\leq \sqrt{\frac{128BC\sqrt{2 n}\int_0^\infty \sqrt{\log N(\epsilon, \Theta, d_3)}d\epsilon +8n\gamma B^2}{n^2} \log \frac{4}{\delta}} \\
          &  + \frac{16\sqrt{2}C}{\sqrt{n}} \int_0^\infty \sqrt{\log N(\epsilon, \Theta, d_3)}d\epsilon  +\frac{16\sqrt{2(N-n)}C}{N} \int_0^\infty \sqrt{\log N(\epsilon, \Theta, d_2)}d\epsilon \\
          &  +\sqrt{\frac{24B^2(N-n)}{N^2}\log \frac{4}{\delta}}+ \frac{4B}{3N} \log \frac{4}{\delta},
  \end{align*}
  where $\gamma=\sup_{\btt\in\Theta}\var (\wt\ell_\btt)/\var (\ell_\btt) \le 1$, and $B$ is the upper bound of the loss function $\ell(f_\btt(\x),y)$.

  \noindent (ii) If additionally $\wh Y_k=Y$ for some $k\in\{1,\dots,K\}$, we have that with probability at least $1-\delta$,
    \begin{align*}
       \calR(\wh \btt^\sada) - &\calR(\btt^*)\leq \frac{16\sqrt{2}C}{\sqrt{N}} \int_0^\infty \sqrt{\log N(\epsilon, \Theta, d_1)}d\epsilon+\sqrt{\frac{24B^2}{N}\log \frac{2}{\delta}}+ \frac{2B}{3N} \log \frac{2}{\delta},
  \end{align*}
  \end{theorem}

  \begin{proof}[Proof of Theorem~\ref{thm:excess_risk_supp}]
    Recall the simplified notations 
    \begin{align*}
        \ell_\btt = \ell(f_\btt(\x),\y) ,~~ 
        \wh \ell_\btt = \frac{N}{N-n}\bm\omega\trans \calL(f_\btt(\x),\wh \y) ,~~
        \wt{\ell}_\btt : = \ell_\btt - \frac{N-n}{N}\wh \ell_\btt,
    \end{align*}
    \begin{align*}
        \ell_{\btt,i} = \ell(f_\btt(\x_i),\y_i) ,~~ 
        \wh \ell_{\btt,i} = \frac{N}{N-n}\bm\omega\trans \calL(f_\btt(\x_i),\wh \y_i) ,~~
        \wt{\ell}_{\btt,i} : = \ell_{\btt,i} -\frac{N-n}{N} \wh \ell_{\btt,i}.
    \end{align*}
      We can rewrite the loss functions as
      \begin{align*}
       \wh \btt^\nv  =  \amin_{\btt\in\Theta} \wh\calR^\nv(\btt)= \amin_{\btt\in\Theta} \frac{1}{n} \sum_{i=1}^n\ell_{\btt,i},
      \end{align*}
      and
      \begin{align*}
          \wh \btt^\sada =\amin_{\btt\in\Theta} \wh\calR^\sada(\btt) = \amin_{\btt\in\Theta} \frac{1}{n} \sum_{i=1}^n\wt{\ell}_{\btt,i} +  \frac{N-n}{N}  \frac{1}{N-n}\sum_{i=n+1}^N \wh \ell_{\btt,i}.
      \end{align*}
      Notice that
      \begin{align*}
          \calR(\wh \btt^\nv) - \calR(\btt^*) &= \calR(\wh \btt^\nv) + \wh\calR^\nv(\btt^*) -\wh\calR^\nv(\btt^*) - \calR(\btt^*)\\
          & \leq \calR(\wh \btt^\nv) + \wh\calR^\nv(\btt^*) -\wh\calR^\nv(\wh \btt^\nv) - \calR(\btt^*)\\
          & \leq  \sup_{\btt \in \Theta}\{\wh\calR^\nv(\btt) - \calR(\btt) \} +\sup_{\btt \in \Theta}\{\calR(\btt) -\wh\calR^\nv(\btt)  \}.
      \end{align*}
      Denote $g(S) =  \sup_{\btt \in  \Theta} \{ \wh\calR^\nv(\btt) - \calR(\btt)  \} $. By the \Cref{ass:complexity}, we know that $\wh\calR^\nv(\btt) - \calR(\btt)$ is a sub-Gaussian process with respect to the metric $\frac{Cd_1}{\sqrt{n}}$. Applying the classical Dudley's entropy integral bound, we have 
      \begin{align*}
          \mE g(S) \leq 8\sqrt{2} \int_0^\infty \sqrt{\log N(\epsilon, \Theta, Cd_1/ \sqrt{n})} d\epsilon &= 8\sqrt{2} \int_0^\infty \sqrt{\log N(\epsilon\sqrt{n}/C, \Theta, d_1)}d\epsilon \\
          &  = \frac{8\sqrt{2}C}{\sqrt{n}} \int_0^\infty \sqrt{\log N(\epsilon, \Theta, d_1)}d\epsilon.
      \end{align*}
      Also, by the \Cref{ass:complexity} and the Talagrand's concentration inequality \citep[Theorem 8.7]{sen2018gentle}, the following holds with probability at least $1-{\delta}/{2}$,
      \begin{align*}
          g(S) \leq \mE[g(S)] + \sqrt{\frac{6B^2}{n}\log \frac{2}{\delta}}+ \frac{B}{3n} \log \frac{2}{\delta}.
      \end{align*}

       Using the same argument, we can similarly bound $\sup_{\btt \in \Theta}\{\calR(\btt) -\wh\calR^\nv(\btt)  \}$. Combining them together, we know it holds with probability at least $1-\delta$,
      \begin{align*}
          \calR(\wh \btt^\nv) - \calR(\btt^*) \leq \frac{16\sqrt{2}C}{\sqrt{n}} \int_0^\infty \sqrt{\log N(\epsilon, \Theta, d_1)}d\epsilon+\sqrt{\frac{24B^2}{n}\log \frac{2}{\delta}}+ \frac{2B}{3n} \log \frac{2}{\delta}.
      \end{align*}
      We will decompose the SADA estimator $\wh\calR^\sada(\btt)$ into two terms and handle each of them using the same method as above,
      \begin{align*}
           \wh\calR^\sada(\btt) = \frac{1}{n} \sum_{i=1}^n\wt{\ell}_{\btt,i} +  \frac{N-n}{N}  \frac{1}{N-n}\sum_{i=n+1}^N \wh \ell_{\btt,i} = \wh\calR_3(\btt)+ \frac{N-n}{N}\wh\calR_2(\btt).
      \end{align*}
      where $\wh\calR_2(\btt) =  {(N-n)}^{-1}\sum_{i=n+1}^N \wh \ell_{\btt,i}$ and $\wh\calR_3(\btt) = {n}^{-1} \sum_{i=1}^n\wt{\ell}_{\btt,i}$.
      By the same argument as before,
      \begin{align*}
          \calR(\wh \btt^\sada) - \calR(\btt^*) &\leq  \sup_{\btt \in \Theta} \left\{\wh\calR^\sada(\btt) - \calR(\btt) \right\} +\sup_{\btt \in \Theta} \left\{\calR(\btt)-\wh\calR^\sada(\btt)  \right\}\\
          & \leq \sup_{\btt \in \Theta} \left\{\wh\calR_3(\btt) - \mE [\wt{\ell}_\btt] \right\} + \sup_{\btt \in \Theta} \left\{\mE [\wt{\ell}_\btt]-\wh\calR_3(\btt)  \right\} \\
          &\quad +\frac{N-n}{N}\sup_{\btt \in \Theta} \left\{\wh\calR_2(\btt) - \mE [\wh \ell_\btt] \right\}+\frac{N-n}{N}\sup_{\btt \in \Theta} \left\{\mE [\wh \ell_\btt] - \wh\calR_2(\btt)  \right\}.
      \end{align*}
      It follows from the exact same argument as $\wh\calR^\nv(\btt)$ that the following holds with probability at least $1-{\delta}/{2}$,
      \begin{align*}
           \sup_{\btt \in \Theta} \left\{\wh\calR_2(\btt) - \mE [\wh \ell_\btt] \right\}+\sup_{\btt \in \Theta} \left\{\mE [\wh \ell_\btt] - \wh\calR_2(\btt)  \right\} &\leq \frac{16\sqrt{2}C}{\sqrt{N-n}} \int_0^\infty \sqrt{\log N(\epsilon, \Theta, d_2)}d\epsilon \\
           & \quad +\sqrt{\frac{24B^2}{N-n}\log \frac{4}{\delta}}+ \frac{2B}{3(N-n)} \log \frac{4}{\delta}.
      \end{align*}
      Denote $f(S): =\sup_{\btt \in \Theta} \left\{\wh\calR_3(\btt) - \mE [\wt{\ell}_\btt] \right\} $. Again, by the Dudley's entropy integral bound, we have
      \begin{align*}
          \mE f(S) \leq 8\sqrt{2} \int_0^\infty \sqrt{\log N(\epsilon, \Theta, Cd_3/\sqrt{n})} d\epsilon \leq \frac{8\sqrt{2}C}{\sqrt{n}} \int_0^\infty \sqrt{\log N(\epsilon, \Theta, d_3)} d\epsilon
      \end{align*}
      By the Talagrand's inequality, it holds with probability $1-{\delta}/{4}$,
      \begin{align*}
           f(S) \leq \mE f(S) + \sqrt{\frac{32BC\sqrt{2 n}\int_0^\infty \sqrt{\log N(\epsilon, \Theta, d_3)}d\epsilon +2n\gamma B^2}{n^2} \log \frac{4}{\delta}} + \frac{B}{3n} \log \frac{4}{\delta}.
      \end{align*}
      We can bound $\sup_{\btt \in \Theta} \left\{\mE [\wt{\ell}_\btt]-\wh\calR_3(\btt)  \right\}$ similarly. Putting all pieces together, it holds with probability at least $1-\delta$,
      \begin{align*}
          \calR(\wh \btt^\sada) - \calR(\btt^*) &\leq \sqrt{\frac{128BC\sqrt{2 n}\int_0^\infty \sqrt{\log N(\epsilon, \Theta, d_3)}d\epsilon +8n\gamma B^2}{n^2} \log \frac{4}{\delta}} \\
          & \quad + \frac{16\sqrt{2}C}{\sqrt{n}} \int_0^\infty \sqrt{\log N(\epsilon, \Theta, d_3)}d\epsilon+\frac{16\sqrt{2(N-n)}C}{N} \int_0^\infty \sqrt{\log N(\epsilon, \Theta, d_2)}d\epsilon \\
          & \quad +\sqrt{\frac{24B^2(N-n)}{N^2}\log \frac{4}{\delta}}+ \frac{4B}{3N} \log \frac{4}{\delta}.
      \end{align*}
       When there exists $k \in \{1,\dots,K\}$ so that $Y = \wh Y_k$, we have $\wh\calR^\sada(\btt) = \frac{1}{N} \sum_{i=1}^N \ell_{\btt,i}$, which essentially has the same form as $\wh\calR^\nv(\btt)$ but replace $n$ with $N$. So following the same logic as before, we have
       \begin{align*}
       \calR(\wh \btt^\sada) - &\calR(\btt^*)\leq \frac{16\sqrt{2}C}{\sqrt{N}} \int_0^\infty \sqrt{\log N(\epsilon, \Theta, d_1)}d\epsilon+\sqrt{\frac{24B^2}{N}\log \frac{2}{\delta}}+ \frac{2B}{3N} \log \frac{2}{\delta}.
  \end{align*}
  \end{proof}

\begin{theorem}
  \label{thm:naive_minimax_supp}
There exist a parameter space $\Theta$, a model $\{f_\btt:\btt\in\Theta\}$, a bounded loss $\ell$, and a class $\calP$ of joint distributions of $(X,Y)$ satisfying Assumption~\ref{ass:complexity} with uniform constants, together with a constant $c>0$, such that for all $n\ge 1$,
\[
\inf_{\widetilde\btt}\sup_{P\in\calP}
\mathbb E_{P^{\otimes n}}\{\calR_P(\widetilde\btt)-\calR_P(\btt_P^*)\}\ge c\,n^{-1/2},
\]
where $\calR_P(\btt)=\mathbb E_P\{\ell(f_\btt(X),Y)\}$, $\btt_P^*\in\arg\min_{\btt\in\Theta}\calR_P(\btt)$, and the infimum is over all measurable estimators based on $n$ i.i.d.\ labeled observations from $P$. Combined with the upper bound in Theorem~\ref{thm:excess_risk_supp}(i), this shows that the naive estimator $\wh\btt^\nv$ attains the minimax rate $n^{-1/2}$ for the labeled-only prediction problem.
\end{theorem}

\begin{proof}[Proof of Theorem~\ref{thm:naive_minimax_supp}]
We exhibit an explicit one-dimensional submodel over which no labeled-only estimator can achieve expected excess risk smaller than order $n^{-1/2}$.

Let $\Theta=[-1,1]\subseteq\mathbb R$, $X\equiv 0$, $f_\theta(0)=\theta$, and $Y\in\{-1,1\}$. Take the bounded continuous loss
\[
\ell(f_\theta(X),Y)=\frac{B}{2}(1-Y\theta),
\]
so that $\ell\in[0,B]$ for all $\theta\in[-1,1]$ and $Y\in\{-1,1\}$. For $v\in\{-1,1\}$ and $0<h\le 1/2$, define $P_{v,h}$ by
\[
P_{v,h}(Y=1)=\frac{1+vh}{2},\qquad P_{v,h}(Y=-1)=\frac{1-vh}{2}.
\]

Take $\calP=\{P_{v,h}:v\in\{-1,1\},\,0<h\le 1/2\}$. The class $\calP$ satisfies Assumption~\ref{ass:complexity} with uniform constants: $\ell\in[0,B]$ gives part~(i); $\ell_\theta-\ell_{\theta'}=(B/2)Y(\theta'-\theta)$ is bounded, hence sub-Gaussian with parameter at most a universal multiple of its standard deviation, so it belongs to $\calF_C$ for a $C$ independent of $v$ and $h$, giving part~(ii); and the entropy integral $\int_0^\infty\sqrt{\log N(\epsilon,\Theta,d_j)}\,d\epsilon$ over the bounded interval $\Theta=[-1,1]$ is finite uniformly, giving part~(iii).

Under $P_{v,h}$, $\mathbb E_{P_{v,h}}[Y]=vh$, so the population risk $\mathcal R_{v,h}(\theta)=B/2-Bvh\theta/2$ is minimized at $\theta_{v,h}^*=v$, and
\[
\mathcal R_{v,h}(\theta)-\mathcal R_{v,h}(\theta_{v,h}^*)=\frac{Bh}{2}(1-v\theta),\qquad \theta\in[-1,1].
\]

Let $\widetilde\btt$ be any estimator based on $n$ i.i.d.\ labeled observations from $P_{v,h}$. Since the submodel is one-dimensional, we identify $\widetilde\btt$ with a scalar $\widetilde\theta\in[-1,1]$ and define the induced test
\[
\widetilde v=
\begin{cases}
1, & \widetilde\theta\ge 0,\\
-1, & \widetilde\theta<0.
\end{cases}
\]
If $\widetilde v\neq v$, then $v\widetilde\theta\le 0$, and hence
\[
\mathcal R_{v,h}(\widetilde\theta)-\mathcal R_{v,h}(\theta_{v,h}^*)=\frac{Bh}{2}(1-v\widetilde\theta)\ge \frac{Bh}{2}.
\]
Thus
\[
\sup_{v\in\{-1,1\}}
\mathbb E_{P_{v,h}^{\otimes n}}
\left[
\mathcal R_{v,h}(\widetilde\theta)
-
\mathcal R_{v,h}(\theta_{v,h}^*)
\right]
\ge
\frac{Bh}{2}
\sup_{v\in\{-1,1\}}
P_{v,h}^{\otimes n}(\widetilde v\neq v).
\]
Taking the infimum over all estimators $\widetilde\btt$ and relaxing to the
infimum over all tests $\phi$ gives
\[
\inf_{\widetilde\btt}
\sup_{v\in\{-1,1\}}
\mathbb E_{P_{v,h}^{\otimes n}}
\left[
\mathcal R_{v,h}(\widetilde\theta)
-
\mathcal R_{v,h}(\theta_{v,h}^*)
\right]
\ge
\frac{Bh}{2}
\inf_{\phi}
\sup_{v\in\{-1,1\}}
P_{v,h}^{\otimes n}(\phi\neq v).
\]
By Le Cam's two-point lemma~\citep[Theorem~2.2(i)]{Tsybakov2009},
\[
\inf_{\phi}
\sup_{v\in\{-1,1\}}
P_{v,h}^{\otimes n}(\phi\neq v)
\ge
\frac12
\left\{
1-
\operatorname{TV}
\left(
P_{+,h}^{\otimes n},
P_{-,h}^{\otimes n}
\right)
\right\}.
\]
A direct calculation gives
\[
D_{\mathrm{KL}}(P_{+,h}\,\|\,P_{-,h})=h\log\frac{1+h}{1-h}\le 4h^2,
\]
for $0<h\le 1/2$, where the last inequality uses $\log\tfrac{1+h}{1-h}\le \tfrac{2h}{1-h}\le 4h$. By tensorization, $D_{\mathrm{KL}}(P_{+,h}^{\otimes n}\,\|\,P_{-,h}^{\otimes n})\le 4nh^2$, and Pinsker's inequality~\citep[Lemma~2.5]{Tsybakov2009} yields
\[
\operatorname{TV}\bigl(P_{+,h}^{\otimes n},P_{-,h}^{\otimes n}\bigr)\le \sqrt{\tfrac12 D_{\mathrm{KL}}(P_{+,h}^{\otimes n}\,\|\,P_{-,h}^{\otimes n})}\le \sqrt{2nh^2}.
\]
Choosing $h=1/(4\sqrt n)$ gives $h\le 1/2$ and
\[
\operatorname{TV}\bigl(P_{+,h}^{\otimes n},P_{-,h}^{\otimes n}\bigr)\le \frac{\sqrt 2}{4}<\frac12.
\]
Consequently,
\[
\inf_{\widetilde\btt}
\sup_{v\in\{-1,1\}}
\mathbb E_{P_{v,h}^{\otimes n}}
\left[
\mathcal R_{v,h}(\widetilde\theta)
-
\mathcal R_{v,h}(\theta_{v,h}^*)
\right]
\ge
\frac{Bh}{2}\cdot \frac12\left(1-\frac12\right)
=
\frac{Bh}{8}
=
\frac{B}{32\sqrt n}.
\]
Since $\{P_{+,h},P_{-,h}\}\subseteq\calP$, this implies
\[
\inf_{\widetilde\btt}
\sup_{P\in\calP}
\mathbb E_{P^{\otimes n}}
\left[
\calR_P(\widetilde\btt)-\calR_P(\btt_P^*)
\right]
\ge \frac{B}{32}\,n^{-1/2},
\]
which completes the proof.
\end{proof}

\section{Additional simulation results}\label{sec_supp:add_simu}
\Cref{tab:supp_bias} and \ref{tab:supp_cp} present the bias and coverage probability of the methods examined in the simulation studies described in Section 5.

\begin{table}[h!]
\caption{Bias of different methods under varying prediction quality $\gamma$.}
\label{tab:supp_bias}
\setlength{\tabcolsep}{3.1pt}
\resizebox{\textwidth}{!}{
\begin{tabular}{@{}ccrrrrrrrrrrr@{}}
\toprule
Method &  &\multicolumn{10}{c}{Bias}\\\midrule
   & & $\gamma=0$  & $\gamma=0.1$ & $\gamma=0.2$ & $\gamma=0.3$ & $\gamma=0.4$ & $\gamma=0.5$ & $\gamma=0.6$ & $\gamma=0.7$ & $\gamma=0.8$ & $\gamma=0.9$ & $\gamma=1$   \\[5pt]  
  Naive & - & 0.007 & 0.007 & -0.001 & -0.002 & -0.001 & 0.006 & -0.004 & -0.009 & 0.002 & 0.009 & -0.003 \\   
  \multirow{2}{*}{PPI} & $\wh{Y}_1$ & 0.005 & 0.009 & -0.005 & -0.002 & -0.001 & 0.004 & 0.001 & -0.004 & 0.000 & 0.002 & 0.003 \\ 
   & $\wh{Y}_2$ & -0.001 & 0.000 & -0.003 & 0.001 & -0.003 & 0.000 & -0.002 & -0.001 & 0.005 & 0.010 & 0.001 \\ 
  \multirow{2}{*}{PPI++} & $\wh{Y}_1$ & 0.007 & 0.007 & -0.001 & -0.003 & -0.002 & 0.005 & 0.000 & -0.006 & 0.001 & 0.003 & 0.000 \\ 
   & $\wh{Y}_2$ & 0.002 & 0.003 & -0.003 & 0.000 & -0.003 & 0.003 & -0.003 & -0.006 & 0.003 & 0.009 & -0.003 \\ 
  SADA & both & 0.002 & 0.002 & -0.003 & -0.000 & -0.003 & 0.003 & 0.000 & -0.005 & 0.001 & 0.003 & 0.000 \\ 
\bottomrule
\end{tabular}
}
\end{table}

\begin{table}[h!]
\caption{Coverage probability of different methods under varying prediction quality $\gamma$.}
\label{tab:supp_cp}
\setlength{\tabcolsep}{3.1pt}
\resizebox{\textwidth}{!}{
\begin{tabular}{@{}ccccccccccccc@{}}
\toprule
Method &  &\multicolumn{10}{c}{Coverage probability}\\\midrule
   & & $\gamma=0$  & $\gamma=0.1$ & $\gamma=0.2$ & $\gamma=0.3$ & $\gamma=0.4$ & $\gamma=0.5$ & $\gamma=0.6$ & $\gamma=0.7$ & $\gamma=0.8$ & $\gamma=0.9$ & $\gamma=1$   \\[5pt]  
   Naive & - & .947 & .940 & .957 & .951 & .943 & .930 & .931 & .953 & .945 & .955 & .947 \\ 
\multirow{2}{*}{PPI} & $\wh{Y}_1$ & .948 & .954 & .957 & .949 & .939 & .952 & .947 & .938 & .950 & .946 & .935 \\ 
   & $\wh{Y}_2$ & .962 & .953 & .942 & .953 & .946 & .937 & .939 & .959 & .951 & .948 & .954 \\ 
  \multirow{2}{*}{PPI++} & $\wh{Y}_1$ & .942 & .939 & .955 & .954 & .931 & .935 & .936 & .941 & .942 & .958 & .935 \\ 
   & $\wh{Y}_2$ & .955 & .944 & .943 & .945 & .944 & .929 & .924 & .950 & .938 & .951 & .945 \\ 
SADA & both & .946 & .932 & .939 & .949 & .938 & .932 & .938 & .943 & .937 & .957 & .924 \\
\bottomrule
\end{tabular}
}
\end{table}

\section{Generating predictions using LLMs}\label{sec:prompt}
In this section, we provide a step-by-step description of how we generate predictions using the large language models like GPT-4o, Llama-3-8B, and DeepSeek in the empirical study presented in \Cref{ssec:exp_politeness}.

The input provided to the LLM consists of four components: a \emph{detail} section that gives natural language descriptions of each data column, a \emph{background} section that provides shared task context for all observations for example what kind of text is being evaluated and how politeness is defined, a \emph{question} section that specifies what the model is asked to predict for every data point and describes the required output format for example a politeness score on a fixed scale., and a set of ten \emph{prompts}, each designed to represent a different tone or inquiry style. As a result, each data point produces ten separate outputs, one for each prompt, and the final prediction is obtained by averaging these outputs.

\begin{tcolorbox}[breakable,colback=gray!10, colframe=black, title=Input prompt for generating responses]
\textbf{Details:} 
The text under review is: \textcolor{magenta}{\{text\}}, and is from the community: \textcolor{magenta}{\{source\}}. Please evaluate its level of politeness based on linguistic features.

\vspace{1em}
\textbf{Background:} 
\textcolor{magenta}{\{backgroud\}}.

\vspace{1em}
\textbf{Question:}
\textcolor{magenta}{\{question\}}.

\vspace{1em}
\textbf{Prompts:} 
\textcolor{magenta}{\{prompt 1\}},..., \textcolor{magenta}{\{prompt 10\}}.
\end{tcolorbox}

Below is an example of the prompts used in the politeness evaluation experiment.

\begin{tcolorbox}[breakable,colback=gray!10, colframe=black, title=An example prompt used in the politeness evaluation experiment]
\textbf{Details:} 
The text under review is: \emph{``Doing a redundant `atoi' after you just did `strtol' is probably the most twisted insult to the proper use of string-to-int conversion functions in C one can come up with. Why do you see the need to ``re-convert'' the value using the broken function `atoi' when you already have it as `val' from a proper function `strtol'?''}, and is from the community: \emph{Stack Overflow}. Please evaluate its level of politeness based on linguistic features.

\vspace{1em}
\textbf{Background:} 
This is a post from an online blog. 
Based on its linguistic features—such as word choice and tone—please evaluate the level of politeness. 
Then, assign a score from 1 (very impolite) to 25 (extremely polite), just like a sommelier would.

\vspace{1em}
\textbf{Question:}
Choose an integer between 1 and 25 that reflects your evaluation. No explanation is needed, and do not give any numbers other than the rating. Your answer must be in JSON format with an integer only, without additional text.

\vspace{1em}
\textbf{Prompts:} 
\begin{itemize}
    \item Given the writing style and overall tone of this blog post, how would you rate its level of politeness?
    \item Taking into account the language used, the word choices, and the sentiment conveyed, what politeness score would you assign?
    \item Imagine you're an expert in online communication. Based on the phrasing and attitude in this post, what number would you give for politeness?
    \item As someone who understands tone and subtext, how polite does this blog post feel to you?
    \item Just like a sommelier tasting wine, how would you score the politeness of this post, based solely on its language?
    \item If you had to assign a politeness score between 1 and 25 to this writing sample, what would it be?
    \item What's your judgment on the tone of this blog post—how polite does it seem on a scale from 1 to 25?
    \item Considering the linguistic subtleties in this post—formality, tone, and word choice—how would you evaluate its politeness?
    \item Reflecting on how this post might come across to a casual reader, what politeness score would you give it?
    \item Evaluate the post as if you were rating its etiquette. What score feels most appropriate?
\end{itemize}
\end{tcolorbox}

\end{document}